\documentclass{article} 

\usepackage{iclr2023_conference,times}

\usepackage{amsmath,amsfonts,bm}

\def\eqref#1{equation~\ref{#1}}

\def\1{\bm{1}}

\DeclareMathAlphabet{\mathsfit}{\encodingdefault}{\sfdefault}{m}{sl}
\SetMathAlphabet{\mathsfit}{bold}{\encodingdefault}{\sfdefault}{bx}{n}

\usepackage{hyperref}
\usepackage{url}
\usepackage{hyperref}
\usepackage[most]{tcolorbox}
\usepackage{wrapfig}
\usepackage{graphicx,xcolor,float}
\usepackage{subfigure}
\usepackage{threeparttable}
\usepackage{algorithm}
\usepackage[noend]{algorithmic}
\usepackage{colortbl}
\usepackage{color}
\usepackage{multirow}
\usepackage{tabularx}
\usepackage{float}
\usepackage{graphicx}
\usepackage{booktabs}
\usepackage{arydshln}
\usepackage{enumitem}
\usepackage{wrapfig}
\usepackage{caption}
\usepackage{graphicx}
\usepackage{makecell}
\usepackage{float} 
\usepackage{makecell}
\usepackage{tabularx}
\usepackage{amssymb,mathrsfs,amsmath}

\usepackage{pifont}
\usepackage{xcolor}
\definecolor{bittersweet}{rgb}{1.0, 0.44, 0.37}
\definecolor{mygreen}{rgb}{0.29, 0.7, 0.48}
\usepackage{pifont}
\definecolor{demphcolor}{RGB}{144,144,144}
\newcommand{\demph}[1]{\textcolor{demphcolor}{#1}}
\definecolor{mygray}{gray}{0.4}
\hypersetup{
    colorlinks=true,
    linkcolor=red,
    citecolor=cyan,
    filecolor=magenta,      
    urlcolor=magenta,
    }

\usepackage{xcolor}

\newcommand{\tabincell}[2]{\begin{tabular}{@{}#1@{}}#2\end{tabular}}

\usepackage[font=small,labelfont=bf]{caption}  
\usepackage{makecell}
\usepackage{tabulary}

\title{The Snowflake Hypothesis: Training Deep GNN with One Node One Receptive field}

\author{
Kun Wang$^{1, \dagger}$,\; 
Guohao Li$^{2, \dagger}$,\; 
Shilong Wang$^{1}$,\; 
Guibin Zhang$^{4}$,\; 
Kai Wang$^{5}$,\; 
Yang You$^{5}$, \\
\textbf{Xiaojiang Peng}$^{6}$,
\textbf{Yuxuan Liang}$^{3,*}$,
\textbf{Yang Wang}$^{1,7,8}$\thanks{Yang Wang and Yuxuan Liang are the corresponding authors, $\dagger$ denotes equal contributions.}
	\\
	$^{1}$University of Science and Technology of China (USTC) \quad \\
    $^{2}$King Abdullah University of Science and Technology (KAUST) \\
    $^{3}$INTR \& DSA Thrust, The Hong Kong University of Science and Technology (Guangzhou) \\
    $^{4}$Tongji University \quad
    $^{5}$National University of Singapore \quad  
    $^{6}$Shenzhen Technology University \\
    $^{7}$School of Software Engineering, USTC \quad $^{8}$School of Data Science, USTC \\
}

\iclrfinalcopy 
\begin{document}

\maketitle

\begin{abstract}
\vspace{-0.6em}
Despite Graph Neural Networks (GNNs) demonstrating considerable promise in graph representation learning tasks, GNNs predominantly face significant issues with over-fitting and over-smoothing as they go deeper as models of computer vision (CV) realm. Given that the potency of numerous CV and language models is attributable to that support reliably training very deep architectures, we conduct a systematic study of deeper GNN research trajectories. Our findings indicate that the current success of deep GNNs primarily stems from (I) the adoption of innovations from CNNs, such as residual/skip connections, or (II) the tailor-made aggregation algorithms like DropEdge. However, these algorithms often lack intrinsic interpretability and indiscriminately treat all nodes within a given layer in a similar manner, thereby failing to capture the nuanced differences among various nodes. In this paper, we introduce \textit{the Snowflake Hypothesis} -- a novel paradigm underpinning the concept of ``one node, one receptive field''. The hypothesis draws inspiration from the unique and individualistic patterns of each snowflake, proposing a corresponding uniqueness in the receptive fields of nodes in the GNNs.

We employ the simplest gradient and node-level cosine distance as guiding principles to regulate the aggregation depth for each node, and conduct comprehensive experiments including: (1) different training scheme; (2) various shallow and deep GNN backbones, especially on deep frameworks such as JKNet, ResGCN, PairNorm \textit{etc.} (3) various numbers of layers (8, 16, 32, 64) on multiple benchmarks (six graphs including dense graphs with millions of nodes); (4) compare with different aggregation strategies. The observational results demonstrate that our framework can serve as a universal operator for a range of tasks, and it displays tremendous potential on deep GNNs.  It can be applied to various GNN frameworks, enhancing its effectiveness when operating in-depth, and guiding the selection of the optimal network depth in an explainable and generalizable way.


\end{abstract}

\vspace{-2mm}
\section{Introduction}
\vspace{-0.5em}
Graph Neural Networks (GNNs)~\citep{kipf2016semi, hamilton2017inductive} have emerged as the leading models for various graph representation learning tasks, including node classification \citep{velickovic2017graph, abu2020n}, link prediction \citep{ zhang2018link, zhang2019inductive}, and graph classification \citep{ying2018hierarchical, gao2019graph}. The prominent performance of GNNs mainly stems from their message passing mechanism \citep{wu2020comprehensive}. Specifically, they iteratively acquire highly-informative representations by aggregating the knowledge from neighbors in the topology \citep{wu2020comprehensive}.

Though promising, over-fitting \citep{rong2019dropedge}, over-smoothing \citep{li2018deeper, chen2020measuring} and vanishing gradients \citep{li2019deepgcns, zhao2019pairnorm} are three long-standing problems in the GNN area, especially when GNNs go deeper as convolutional neural networks (CNNs) \citep{he2016deep}. Consequently, when training an over-parameterized GNN on a small graph or utilizing a deep GNN for graph modeling, we often end up with collapsed weights or indistinguishable node representations~\citep{chen2020measuring}. Therefore, training 2-to-4-layer GNNs is not a foreign phenomenon in the graph realm and most state-of-the-art GNNs are no deeper than 4 layers \citep{sun2019adagcn}. Nonetheless, scrutinizing the brilliant achievements on many computer vision tasks, which can be primarily attributed to the consistent and effective training of deep networks. \citep{he2016deep, huang2017densely}. Graph representation learning eagerly calls for the utilization of deeper GNNs, particularly when dealing with large-scale graphs characterized by dense connections.

Most recently, several works have shown the feasibility of training GNNs with increasing depth. We can summarize the existing approaches into two categories: The first category involves prudently inheriting innovations from CNNs, such as residual/dense connections \citep{li2019deepgcns, sun2019adagcn, xu2018representation, li2021deepgcns, xu2018representation, chen2020simple, xu2021optimization}, which have proven to be universally applicable and practical. For instance, JKNet \citep{xu2018representation} adopts skip connections to fuse the output of each layer to maintain the discrepancies among different nodes. GCNII \citep{chen2020simple} and ResGCN \citep{li2019deepgcns} employ residual connections to carry the information from the previous layer to avoid the aforementioned issues. Another category is to combine various deep aggregation strategies with shallow neural networks \citep{wu2019simplifying, chien2020adaptive, liu2020towards, zou2019layer, rong2019dropedge, gasteiger2019diffusion}. For example, GDC \citep{gasteiger2019diffusion} generalizes Personalized PageRank into a graph diffusion process. DropEdge \citep{rong2019dropedge} resorts to a random edge-dropping strategy to implicitly increase graph diversity and reduce message passing.

However, although CNN inheritances such as residual/skip connections can partially alleviate the over-smoothing problem, these modifications fail to effectively explore the relationship between aggregation strategies and network depth. Incorporating residuals into layers with suboptimal output may inadvertently propagate detrimental information to subsequent aggregation layers. Within the second category, the majority of existing deep aggregation strategies attempt to sample a subset of neighboring nodes around the central node to implicitly enhance data diversity and prevent over-smoothing. Unfortunately, the cumbersome and particular designs make GNN models neither simple nor practical, lacking the ability to scale on other training strategies and specific datasets.

In this paper, we hypothesize that each node within a graph should possess its unique receptive field. This can be actualized through the process of element-level adjacency matrix pruning. Such a procedure enables an ``early stopping'' feature in node aggregation in terms of depth, which not only amplifies interpretability but also aids in mitigating the over-smoothing issue. Drawing from the extensive experimental observations, we first introduce \textit{the Snowflake Hypothesis}.

\textbf{The Snowflake Hypothesis (SnoH).} When fitting an abstracted graph ${\cal G} = \left( {{\cal V},{\cal E}} \right)$ with vertices set $\cal V$ and edges $\cal E$ from the natural world using GNN models (GNNs aim to learn a representation vector of a node or an entire graph based on the adjacency matrix $A \in \mathbb{R}^{|\mathcal{V}| \times |\mathcal{V}|}$ and  node features $X \in \mathbb{R}^{|\mathcal{V}| \times F}$), by utilizing a prune-specific adjacency matrix at the element level, we can uncover the distinctive receptive fields that each node ought to aggregate, akin to the uniqueness and unique patterns observed in each snowflake. This enables us to overcome the over-smoothing problem and train deeper GNNs.

We utilize the simplest gradient (version 1) and node-level cosine distance (version 2) as the guiding principles to control the aggregation depth for each node, and we adopt extensive experiments, including: 1) \textbf{Different training algorithms} such as a pre-training scheme \citep{liu2023pre}, iterative pruning \citep{frankle2018lottery, chen2021unified}, and re-initialization. We found that compared to popular pruning algorithms, our approach not only allowed us to leverage the benefits of pruning but also facilitated better network convergence. 2) \textbf{Integration with mainstream deep GNN models}, such as ResGCN \citep{li2021deepgcns}, JKNet \citep{xu2018representation}, and GCNII \citep{chen2020simple}. This integration enabled us to better assist the network in improving performance and increasing depth. 3) \textbf{Comparing our algorithm with the DropEdge or graph lottery ticket \citep{chen2020lottery} methods}, we observe that our algorithm possesses better interpretability and generalizability. Our algorithm can also be conceptualized as a data augmenter and a message passing reducer, providing a new paradigm to benefit many graph pruning applications.


\textbf{Identifying unique snowflake.} We identify a unique snowflake by employing layer-wise adjacency matrix pruning and node-level cosine distance discrimination. Taking a 2-layer vanilla GCN \citep{velickovic2017graph} as example, assuming that the trainable parameters ${\rm{\Theta }} = \{ {{{\rm{\Theta }}^{\left( 0 \right)}},{{\rm{\Theta }}^{\left( 1 \right)}}} \}$ for node classification as:
\begin{equation}\label{eq:GNN}
\vspace{-0.2em}
\nonumber
\mathcal{Z} = {\rm{Softmax}}\left( {\hat{A}\sigma\left( {\hat{A}X\Theta^{(0)}} \right)\Theta^{(1)}} \right) \text{,\quad loss function: } 
\mathcal{L}\left( {\mathcal{G},\Theta} \right) =  - \mathop \sum \nolimits_{{v_i} \in \mathcal{V}_{l}} {y_i}{\log}\left( {{z_i}} \right) 
\vspace{-0.2em}
\end{equation}
where ${\cal Z}$ is the model predictions, $\sigma \left(  \cdot  \right)$ denotes an activation function, $\hat A = {\tilde S^{ - \frac{1}{2}}}\left( {A + I} \right){\tilde S^{ - \frac{1}{2}}}$ is the normalized adjacency matrix with self-loops and $\tilde{S}$ is the degree matrix of $A + I$. We minimize the cross-entropy loss $\mathcal{L}\left( {\mathcal{G},\Theta} \right)$ over all labelled nodes $\mathcal{V}_l \subset \mathcal{V}$, where ${y_i}$ and ${z_i}$ represents the label and prediction of node ${v_i}$, respectively. We present the first versions (v1) of SnoH as follows:

\vspace{-2mm}
\begin{itemize}[leftmargin=*]
    
    \item[1] Randomly initialize a graph neural network (GNN) $f\left( {{\cal G},{{\rm{\Theta }}_0}} \right)$ for training graph ${\cal G}$.

    \item[2] Train the GNN (total $D$ layer) for $k$ iterations, compute the absolute gradient of each elements in the outermost adjacency matrix (note ${A_{\left( D \right)}}$). Remove the $p\%$ elements with the smallest gradients in the adjacency matrix.      
    
    \item[3] Repeat step 2 by computing ${A_{\left( D-1 \right)}}$ and assigning the index of zero elements in ${A_{\left( D-1 \right)}}$ to ${A_{\left( D \right)}}$, thereby setting corresponding positions of the next layer's adjacency matrix to zero.
    
    \item[4] Repeat steps 2-3 iteratively, removing the corresponding elements from the adjacency matrix and assigning their zero element positions to all deeper layers.

\end{itemize}
\vspace{-0.5em}

By utilizing the simplest gradient guidance which can indicate potentially promising elements \citep{Le2020GoingDW, blalock2020state}, we can realize the Snowflake Hypothesis, enabling each node to have its unique aggregation depth and receptive field size. For ease of understanding, we showcase the detailed training implications in Appendix \ref{Snov2}. However, the task of calculating gradients for the adjacency matrix presents a substantial challenge. Specifically, for larger graphs with millions of nodes such as Ogbn-product, which possesses 61,859,140 edges \citep{hu2020open}. Blindly calculating the gradient for each edge is both difficult and imprudent. The computation of a massive number of parameters makes it extremely challenging to incorporate this algorithm into the deeper GNN framework. To address this, we present SnoHv2, where we make a simple modification to focus on node representations: 

\vspace{-2mm}
\begin{itemize}[leftmargin=*]
    
    \item[1] Randomly initialize a graph neural network (GNN) $f\left( {{\cal G},{{\rm{\Theta }}_0}} \right)$ for training graph ${\cal G}$.

    \item[2] Train $k$ iterations, calculate the cosine distance $D(Z^{\left( l \right)}, {\cal T}( {Z^{\left( l \right)}}) ) = 1 - \frac{Z^{\left( l \right)} \cdot {\cal T}( {Z^{\left( l \right)}})}{||Z^{\left( l \right)}||_2 \cdot ||{\cal T}( {Z^{\left( l \right)}})||_2}$ between the representation of each node in the GNN and its aggregated representation of the surrounding nodes at each layer, $Z^{\left( l \right)}$ denotes the representation of the $i$-th node at layer $l$ before aggregation by the adjacency matrix, while $\cal T$ symbolizes the representation after aggregation. In the context of GCN, $Z^{\left( l \right)}$ is defined as $H^{\left( l \right)} \cdot {W^{\left( l \right)}} $, and ${\cal T}( {Z^{\left( l \right)}})$ is represented as $ A^{(l)} H^{\left( l \right)} {W^{\left( l \right)}} $, where $H^{\left( l \right)}$ is the feature embedding at $l$ layer.

    \item[3] Compute the nodes whose cosine distance \citep{Le2020GoingDW, blalock2020state} is below the $p\%$ of the distance at the initial layer, and remove the element for node aggregation corresponding to these nodes. As an example, for the $i$-th node, this equates to pruning all elements in the $i$-th row of the adjacency matrix (excluding self-loops, which are not pruned).
    
    \item[4] Once the cosine distance of the $i$-th nodes at the $r$-th layer falls below $p\%$, all elements in the $i$-th row of the adjacency matrices at all deeper layers are pruned.

\end{itemize}
\vspace{-0.5em}

The motivation behind SnoHv2 is quite straightforward: \textit{As the depth of the GNN (Graph Neural Network) increases, the issue of over-smoothing becomes more severe. Representations of neighboring nodes tend to converge, which in turn leads to the network losing its discriminative capacity. Implementing early stopping in terms of depth can aid in restoring the expressiveness of the nodes.}

\vspace{-1em}
\section{Implementations \& Contributions}
\vspace{-0.5em}

In our paper, we aim to test various training schemes, backbones, and datasets to explore the effectiveness of SnoH. We also integrate our approach with state-of-the-art deep GNN frameworks and compare it with similar popular algorithms to further illustrate the scalability and generality of our algorithm and hypothesis. 

\textbf{Training schemes.} We select three training methods to explore the performance of our algorithm and the benefits of combining our algorithm with mainstream training approaches: (1) \textbf{SnoHv1/v2(O)}: we adopt the original \textit{hierarchical one-shot adjacency pruning} approach. (2) \textbf{SnoHv1/v2(IP)}: As our work can be regarded as a graph pruning method, we have opted for the widely recognized iterative pruning (IP) strategy \citep{frankle2018lottery} within the framework of unified graph sparsification (UGS) \citep{chen2020lottery}. (3) \textbf{SnoHv1/v2(ReI)}: Due to the challenge of determining whether the model has converged during pruning, after pruning an adjacency matrix, we fix it and \textit{reinitialize} the GNN for the next training phase. We place the details in Appendix \ref{training}. 

\textbf{Datasets \& Backbones.} In this paper, we utilize six graph benchmarks to evaluate the performance of our hypothesis. Specifically, we select three widely-used small graphs, namely Core, citepseer, and PubMed \citep{kipf2016semi}, for node classification. Additionally, to assess the scalability of our proposal, we incorporate three large-scale graphs known as Ogbn-arxiv, Ogbn-proteins and Ogbn-products \citep{hu2020open}. For all selected datasets, we compare our framework with different baseline settings under the same network configurations. We adopt GCN \citep{kipf2016semi}, GIN \citep{xu2018powerful} and GAT \citep{velivckovic2017graph} as shallow GNNs backbones. Further, we take deep ResGCNs \citep{li2020deepergcn}, JkNet \citep{xu2018representation} and PairNorm \citep{zhao2019pairnorm} as deep backbones. To evaluate our framework on the graph classification task, DropEdge \citep{rong2019dropedge} and UGS \citep{chen2020lottery} are leveraged as the comparison algorithms. More details about experimental settings can be found in Appendix \ref{dataset}.  

\textbf{Contributions.} We summarize our contributions as three folds:
\vspace{-0.6em}
\begin{itemize}[leftmargin=*]
   \item We propose a node ``early stopping'' technique based on edge pruning to help better combat the issue of over-smoothing and over-fitting. Based on extensive observational results, we put forth ``The Snowflake Hypothesis'' -- ``one node one receptive field'', which is inspired by the notion that each snowflake is unique and possesses its own pattern. Likewise, each node in GNNs should have its own receptive field, reflecting its unique characteristics.
   \vspace{-0.1em}

   \item Our algorithm inherently possesses explainability and, while inheriting the advantages of the pruning algorithms (accelerating inference time and reducing storage overhead), it can also benefit the current graph pruning algorithms. More importantly, our algorithm is simple and convenient. Compared to the design of complex aggregation strategies, our framework does not introduce any additional information (\emph{e.g.}, learnable parameters), which can be easily scaled up to deep GNNs. 
   \vspace{-0.1em}
   
   \item We conduct comprehensive experiments, spanning an array of training algorithms, integration with various backbone architectures, and comparisons with DropEdge/UGS frameworks, across multiple graph benchmarks. Our findings indicate that SnoHv1/v2 consistently delivers standout performance, even in instances where the adjacency matrix is notably sparse. These results underscore our initial hypothesis: certain nodes necessitate early termination in their depth progression. 
\end{itemize}

\vspace{-0.6em}
\textbf{Prior work.} Our work contributes to the domain of graph pruning algorithms, aligning with the research trajectories of graph sampling \citep{chen2018fastgcn, eden2018provable, chen2021unified} and graph pooling \citep{ranjan2020asap, zhang2021hierarchical, ying2018hierarchical}. In particular, our algorithm shares significant parallels with the prevalent graph lottery ticket pruning algorithm \citep{chen2020lottery}, striving to replicate the performance of an original, unpruned graph by means of iterative pruning. Moreover, we aim to address the over-smoothing and over-fitting problems that may surface during the training of deep GNNs \citep{li2018deeper, chen2020measuring}. Current methodologies in training deep GNNs principally concentrate on two areas: (1) incorporating components such as residual/skip connections from the architecture of CNNs \citep{li2019deepgcns, sun2019adagcn, xu2018representation} , and (2) crafting a diverse array of aggregation strategies \citep{wu2019simplifying, chien2020adaptive, liu2020towards, zou2019layer}. These focal areas also form the bedrock of our proposed research framework. We refer detailed discussions in the appendix \ref{related work}.

\vspace{-0.5em}
\section{Identifying the unique snowflakes in small graphs} \label{sec3}
\vspace{-0.5em}

In this section, we meticulously examine and conduct numerous experiments to validate our hypothesis on several small graphs, namely Cora, citepseer, and PubMed. We choose \textbf{SnoHv1/v2(O)} as the benchmark training scheme. The experimental settings are placed in Appendix \ref{small}.

\begin{table}[t] \small
\caption{Performance comparisons on 8, 16, 32 layer settings using SnoHv1/v2 across three small graphs, all experimental results are the average of \textbf{five runs} and the \textcolor{red}{red font} indicates the optimal value in a set of results. }\label{tab:results1}
\vspace{-1.0em}
\setlength{\tabcolsep}{3pt}
\begin{center}
\def \arraystretch{0.98}
 \begin{tabular}{ccccccccccccc} 
 \toprule
    \multirow{2}{*}{\bf Backbone}  & \multicolumn{2}{c}{\bf 8 layers} & \multicolumn{2}{c}{\bf 16 layers}  &  \multicolumn{2}{c}{\bf 32 layers}  &  \multicolumn{2}{c}{\bf 64 layers}  \\ 
    \cmidrule(l){2-3} \cmidrule(l){4-5} \cmidrule(l){6-7} \cmidrule(l){8-9}  
    
     & \scriptsize \bf Original  
     & \scriptsize \bf SnoHv1/v2 & \scriptsize \bf Original   & \scriptsize \bf SnoHv1/v2
     
     & \scriptsize \bf Original 
     & \scriptsize \bf SnoHv1/v2  

     & \scriptsize \bf Original 
     & \scriptsize \bf SnoHv1/v2    \\ 
     
     \midrule
     
       \multicolumn{9}{l}{\scriptsize{  \demph{ \it{Train scheme: SnoHv1/v2(O), Dataset: Cora, 2-layer performance: GCN without BN = 85.37 } }} }\\
       
        GCN    & 85.11   & 85.17/\textcolor{red}{85.68}    & 83.75  & 83.87/\textcolor{red}{84.19}  & 80.33   & 81.10/\textcolor{red}{83.09} & 66.11  & 68.45 /\textcolor{red}{72.88}  \\
        
        ResGCN    & 85.31   & 85.37/\textcolor{red}{86.11}    & 85.75  & 85.99/\textcolor{red}{86.52} & 86.27  & 86.33/\textcolor{red}{86.64} & 85.21  & 85.24/\textcolor{red}{85.90}  \\

        JKNet    & 86.33   & \textcolor{red}{87.01}/86.53    & 86.28  & 86.17/\textcolor{red}{87.08}     & 87.20   & 87.31/\textcolor{red}{88.86} & 84.84  & 85.19/\textcolor{red}{85.96} \\ 
        PairNorm    & 83.66 & 82.11/\textcolor{red}{85.90}  & 80.29 &  80.44/\textcolor{red}{83.25} & 78.66 & 79.34/\textcolor{red}{83.16} & 74.12 & 74.19/\textcolor{red}{78.60} \\ 
    
    \midrule
           \multicolumn{9}{l}{\scriptsize{  \demph{ \it{Train scheme: SnoHv1/v2(O),  Dataset: citepseer, 2-layer performance: GCN without BN = 72.44 } }} }\\
       
        GCN    & 72.39   & 72.41/\textcolor{red}{73.24}    & 71.28 & 72.10/\textcolor{red}{72.33} & 68.99   &69.21/\textcolor{red}{69.89} & 44.37  & 45.12/\textcolor{red}{46.65} \\
        ResGCN    & 72.11  & 72.07/\textcolor{red}{72.23}   & \textcolor{red}{72.40}  & 71.91/71.91 & 72.43  & 72.44/\textcolor{red}{73.53} & 71.65  & 72.10/\textcolor{red}{72.94} 
        
        \\
        JKNet    & \textcolor{red}{71.77}  & 71.50/71.47    & 70.72  & 70.60/\textcolor{red}{71.47}    & 70.12   & 70.01/\textcolor{red}{ 72.67} & 69.92   & 70.09/\textcolor{red}{71.55} \\ 
        PairNorm     & 72.88 & 72.34/\textcolor{red}{73.84}  & 73.91 &  73.95/\textcolor{red}{74.58} & 73.36 & 73.05/\textcolor{red}{73.92} & 70.88 & 70.85/\textcolor{red}{72.99} \\ 
    
    \midrule
           \multicolumn{9}{l}{\scriptsize{  \demph{ \it{Train scheme: SnoHv1/v2(O), Dataset: PubMed, 2-layer performance: GCN without BN = 86.50 } }} }\\
       
        GCN    & 86.41   & 86.50/\textcolor{red}{86.56}    & 84.77  & 84.74/\textcolor{red}{85.79} & 83.76  & 83.77/\textcolor{red}{84.06}  & 77.29 & 78.15/\textcolor{red}{78.99} \\
        ResGCN    & 87.45   & 87.50/\textcolor{red}{87.84}    & 87.73  & 87.47/\textcolor{red}{88.33} & 87.66  & 87.33/\textcolor{red}{88.49}  &  87.01  & 86.03/\textcolor{red}{88.11} \\
        
        JKNet    & 88.20   & 88.31/\textcolor{red}{88.51}    & 87.32  & 87.62/\textcolor{red}{87.95}   & 88.81  & 88.75/\textcolor{red}{88.99} & 87.25  & 86.98/\textcolor{red}{87.93} \\ 
        PairNorm   & 87.63   & 87.50/\textcolor{red}{88.68}    & 87.92  & 87.74/\textcolor{red}{88.60} & 87.07  & 87.24/\textcolor{red}{87.69}  & 85.41 & 85.48/\textcolor{red}{87.06} \\ 
    
    \midrule

\end{tabular}
\end{center}
\vspace{-6mm}
\end{table}

\begin{figure*}[h]
  \centering
  \includegraphics[width=0.95\linewidth]{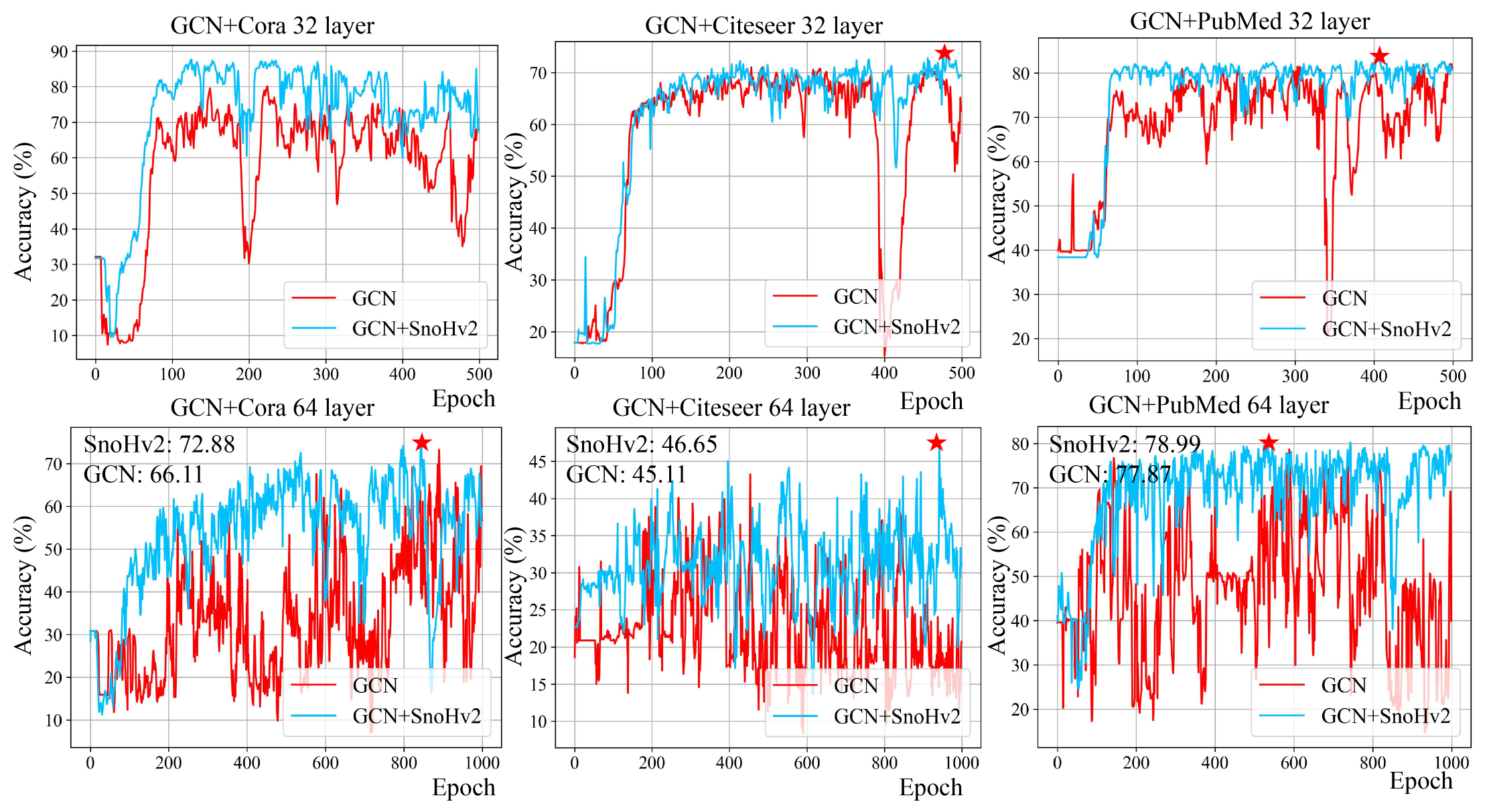}
  \vspace{-0.5em}
  \caption{Performance comparisons on 32, 64 layer settings using SnoHv2 across three small graphs.}
  \label{fig:mainmodel}
  \vspace{-7pt}
\end{figure*}

As depicted in Tab \ref{tab:results1}, under conditions of high sparsity (especially with deep adjacency matrices where sparsity is considerable), SnoHv1 can achieve results approximating those of the original baseline. This observation indicates that implementing early stopping for certain nodes in terms of depth does not compromise the overall performance of the model. Upon transitioning to the more robust SnoHv2 version, we notice a performance enhancement in our model. This further suggests that early stopping in depth may help overcome the over-smoothing phenomenon. As frameworks like ResGCN and JKNet are specifically designed for deep GNNs, we have not presented results for shallow layers. Here, we independently document the results of SnoHv2 for shallow layers. In the case of a 2-layer GCN on Cora, we observe a score of 86.08\% (baseline 85.79\%), on citepseer, it's 73.81\% (baseline 73.58\%), and on PubMed, it's 88.54\% (baseline 88.65\%). We find that, even in shallow GCN, implementing ``early stopping'' for certain nodes in depth could enhance performance (0.29 on Cora and 0.23 on citepseer). With regard to PubMed, we argue that due to the relative largeness, even after two layers of aggregation, better representations may not have been learned. All nodes may require a deeper receptive field, which aligns with the phenomenon observed in the table where extending the depth to between 8-32 layers leads to a performance boost after pruning.

Interestingly, when the depth of the GCN reaches 32/64 layers, SnoHv2 shows a stronger performance boost. Specifically, under the experimental setup of a 64-layer GCN + SnoHv2, improvements of 6.77\%, 1.66\%, and 1.12\% were achieved on the Cora, citepseer, and PubMed, respectively. These astonishing results clearly illustrate the effectiveness of our algorithm. In Tab \ref{tab:32S} in Appendix \ref{small}, we present the sparsity under different datasets. We found that as the network deepens, both node sparsity and edge sparsity are decreasing. At the lower level with high sparsity (approximately $17\%\sim32\%$), some nodes and edges were pruned, which in fact improved the model's performance. This validates the contribution of reducing the receptive field to performance enhancement. In Table \ref{tab:cora64}, we can observe the sparsity of the 64-layer on Cora, which can reach 6.57\% node sparsity and 15.26\% edge sparsity in deeper layer. This further corroborates the notion that many nodes in deep networks do not require such a large receptive field.

\begin{figure*}[h]
  \centering
  \includegraphics[width=0.95\linewidth]{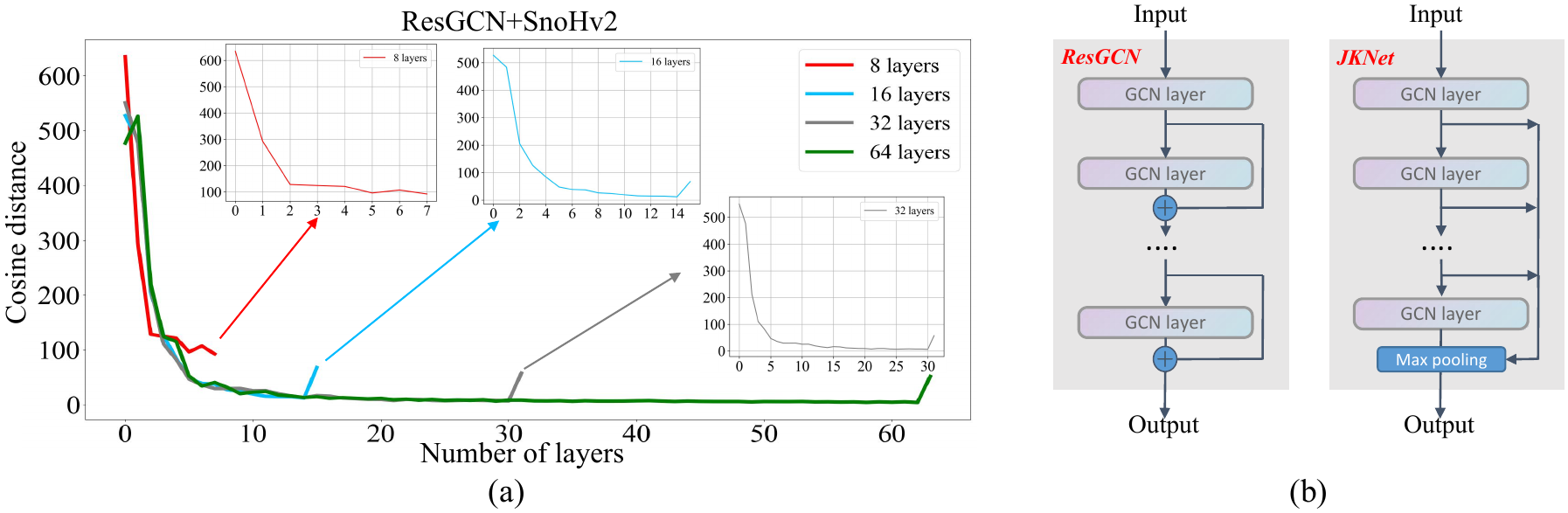}
  \vspace{-0.4em}
  \caption{(a) Cosine distances on ResGCN+SnoHv2 of Cora across various layers settings. (b) ResGCN and JKNet structure overviews.}
  \label{fig:arti}
  \vspace{-7pt}
\end{figure*}

\textbf{How does SnoH help deep GNNs?} We have found that when our framework is integrated, deep GNNs such as ResGCN, JKNet and PairNorm can benefit from our design. We summarize our observations as follows. \textbf{Obs 1.} In deep architectures (32, 64 layers), SnoH brings more significant improvements to ResGCN than in shallow architectures. For instance, we recorded the ResGCN performance as 85.21\%, 71.65\%, 87.01\% on Cora, citepseer, and PubMed at 64 layers. When combined with SnoHv2, the performance is 85.90\%, 72.94\%, 88.11\% (see Fig \ref{fig:res} in the appendix \ref{small}). \textbf{Obs 2.} Taking ResGCN and JKNet as examples (Fig \ref{fig:arti} (b)), we are surprised to find that the sparsity of each layer can help us determine the specific depth that the GCN architecture should retain. For instance, we found that for the combination of JKNet+Cora, the edge sparsity decays to zero after the 13th layer. This indirectly indicates that the aggregation output of the adjacency matrix after the 16th layer no longer contributes to the model, and similar phenomena occur under different combination designs, such as ResGCN+citepseer no more than 26 layers.\textbf{Obs 3.} We computed the cosine distance for each layer of ResGCN+SnoHv2 settings with various depths of GNN (Fig \ref{fig:arti} (a)) on Cora. We observed a gradual reduction in cosine distance as the depth increased, providing further evidence of the existence of the over-smoothing issue, which causes nodes' representations to converge as the network deepens, thus hindering predictive performance. To overcome the over-smoothing problem and enhance interpretability, we applied pruning to stop nodes from growing deeper in the network. Additional results are presented in Appendix \ref{small} Fig  \ref{fig:cosine} and \ref{fig:JKNet}.

\textbf{Compare with pruning algorithm.} Interestingly, our algorithm can be understood as an adjacency matrix pruning algorithm. We chose the current mainstream graph pruning algorithm UGS \citep{chen2020lottery} and random pruning for comparison with the performance of the universal pruning algorithm.  To keep the experimental settings consistent, we removed the part of UGS that targets weight pruning.   And we controlled the iterative pruning rates at 5\%, 10\%, and 20\% and pruning 5 times. In order to make a better comparison, we observed the pruning rates when discovering tickets in the lottery ticket scenario, and recorded the pruning rates of SnoHv2 when it get best performance for comparison. We showcase the comprehensive results in Tab \ref{UGSapp} (Appendix \ref{small}) and we list observations: \textbf{Obs 1.} Our model can achieve better performance than random pruning and UGS, even under higher sparsity levels. This further validates our performance in deeper networks, providing assurance for our algorithm's scalability on large datasets. \textbf{Obs 2.} Compared to the UGS, our method has better interpretability. UGS maintains the same sparsity level for each layer's adjacency matrix, which may lead to the loss of many nodes in shallow training. This is unreasonable as early nodes should aggregate essential information, ensuring they can learn better representations. Our experiments further confirm that the receptive field should gradually increase.

  \vspace{-0.6em}
\begin{table}[h] \footnotesize

\setlength{\tabcolsep}{4pt}
  \caption{Comparison performances of SnoHv2 with UGS and random pruning (RP). Here IPR denotes iterative pruning rate and we set number of layers as 8. We use GCN backbone and set early stopping threshold of cosine distance as $\rho$ (Detailed descriptions in Appendix \ref{small}).} \label{UGS}
  \vspace{-0.6em}
  \centering
  \begin{tabular}{ccccccc}
    \toprule
     \textbf{Dataset}   &   \textbf{RP} &  \textbf{UGS(IPR=5\%)}  &  \textbf{UGS(IPR=10\%)}  & \textbf{UGS(IPR=20\%)} &  \textbf{SnoHv2} &  \textbf{GCN}  \\
    \midrule
    Cora (L=8) & $69.60$ & $73.64$ & $66.01$ & $53.29$ & \cellcolor{gray!30}${85.68}$  & ${85.11}$ \\
      citepseer (L=8) & $45.50$ & $65.80$  & ${51.50}$ & ${43.10}$ & \cellcolor{gray!30}${73.24}$ & ${72.39}$   \\
      PubMed (L=8) & $77.82$ & $84.33$   & $80.91$ & $71.05$ &\cellcolor{gray!30} ${86.56}$ & ${86.41}$   \\ 
    \bottomrule
  \end{tabular}
  \vspace{-1em}
\end{table}

We also migrated potential training strategies in pruning to SnoHv1 (since SnoHv2 fundamentally determines the early stopping depth of each node by similarity, different training strategies don't have much significance in SnoHv2, while the training process of SnoHv1 and the pruning process are very similar). We display the results in Appendix \ref{small}. As can be easily seen, different training strategies do not significantly improve SnoHv1's results. However, in practice, iterative pruning and re-initialization strategies can bring about severe efficiency problems (Even $D\times$ on the training burden of re-initialization). Therefore, we adopt a one-shot pruning strategy as the preferred strategy for SnoHv1. Unless specified otherwise, the performance we present is that of SnoHv1(O).
\vspace{-0.6em}

\textbf{Compare with other edge drop strategy (DropEdge).} Another related approach can be understood as the edge drop strategy, where DropEdge \citep{rong2019dropedge} shares similarities with SnoH. Although DropEdge can improve performance through implicit data augmentation, it lacks interpretability in its aggregation strategy. In fact, it should not continue aggregation after halting the information aggregation for a certain node at a higher layer. It is worth noting that since DropEdge involves temporarily increasing data samples during training, it can be easily combined with SnoHv2. We present the results in Tab \ref{tab:difference} and list observations. 

\textbf{Generalizability on different backbones.} To validate the generalization capability of our algorithm across different backbones, we further selected popular GNN frameworks GIN \citep{xu2018powerful} and GAT \citep{velivckovic2017graph} as the backbones for the generalization evaluation. We controlled the parameter $\rho$ at values of 0.2, 0.1, and 0.05 under experiments with network depths of 8, 16, and 32 layers, respectively, while recording the results.

\begin{minipage}[!b]{0.36\linewidth}
    \centering
		\renewcommand{\arraystretch}{0.8}
    \scriptsize
    \tabcolsep=0.85mm
 \begin{tabular}{c|c|ccc}
    \toprule
    & \multirow{2}{*}{\tabincell{c}{Total \\ layers}} & \multicolumn{3}{c}{Graphs} \\
    & & Cora & citepseer & PubMed \\
    \midrule
    \tabincell{c}{DropEdge} & \tabincell{c}{8 \\ 16 \\ 32} & \tabincell{c}{86.98 \\ 84.01 \\ 80.81} & \tabincell{c}{74.57 \\ 73.17 \\ 71.77} & \tabincell{c}{86.91 \\ 86.35 \\ 82.23} \\
    \midrule
    \tabincell{c}{DropEdge +\\ SnoHv2 }  & \tabincell{c}{8 \\ 16 \\ 32} & \tabincell{c}{81.70 \\ 80.32 \\ 76.64} & \tabincell{c}{72.97 \\ 72.12 \\ 65.56} & \tabincell{c}{87.25 \textcolor{red}{$\uparrow$} \\ 86.99 \textcolor{red}{$\uparrow$} \\ 87.19 \textcolor{red}{$\uparrow$} } \\
    \bottomrule
  \end{tabular}
    \makeatletter\def\@captype{table}\makeatother\caption{Comparison between DropEdge and DropEdge+SnoHv2.
}\label{tab:difference}
    \end{minipage}
     \hspace{0.1em}
    \begin{minipage}[!b]{0.62\linewidth}
    \centering  
    \includegraphics[width=.95\linewidth]{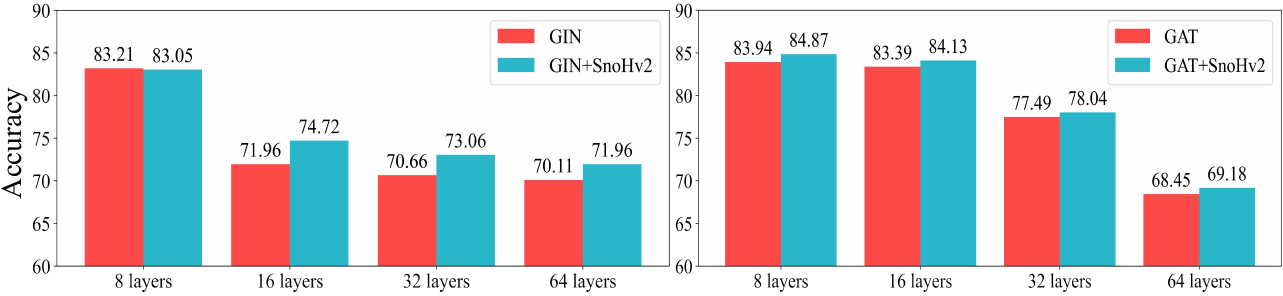}
  \makeatletter\def\@captype{figure}\makeatother\caption{Experiment results on different graph backbones (GIN, GAT) across Cora dataset. The additional results on citepseer and PubMed can be found in Appendix \ref{genapp}.} \label{generability}
\end{minipage}

We observe that our hypothesis remains viable when applied to GIN and GAT. When combined with SnoHv2, these backbones still demonstrate improved performance in deeper layers. Specifically, on a 16-layer GIN, we achieve a gain of 2.46\%, and on a 64-layer GAT, we achieve a gain of 0.73\%. These results further support the generalization capability of our algorithm.

\vspace{-0.6em}
\section{SnoHv1/v2 on large-scale graphs}\label{sec4}
\vspace{-0.6em}
In this section, we thoroughly assess our hypothesis on large graphs, employing three benchmark datasets: Ogbn-Arxiv, Ogbn-Proteins, and Ogbn-Product. The dataset splits adhere to the guidelines provided by \citep{hu2020open}. For Ogbn-ArXiv, our training data comprises papers published until 2017, validation data encompasses papers published in 2018, and testing data includes papers published since 2019. Regarding Ogbn-Proteins, we partition proteins into training, validation, and test sets based on their species. For the Product dataset, we adopt sales ranking as the criterion to divide nodes into training, validation, and test sets. Specifically, we assign the top 8\% of products to the training set, the next top 2\% to the validation set, and the remainder to the test set.
\vspace{-0.6em}
\subsection{SnoHv1/v2 on Citation Network}
\vspace{-0.6em}

\begin{figure*}[h]
  \centering
  \includegraphics[width=0.95\linewidth]{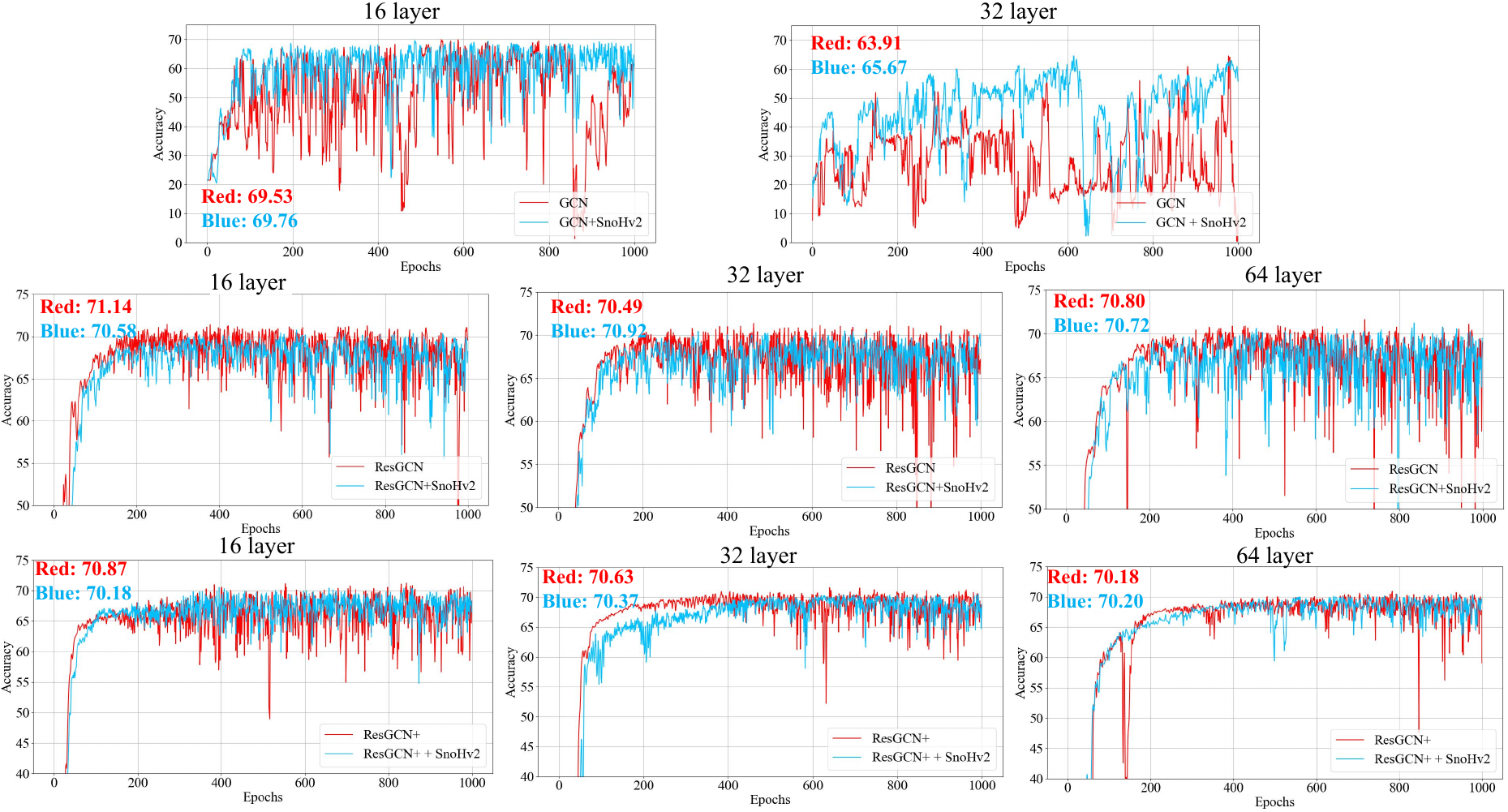}
  \vspace{-0.5em}
  \caption{Performance comparisons on 16, 32, 64 layer settings using SnoHv2 across GCN, ResGCN and ResGCN+ on Ogbn-Arxiv.}
  \label{fig:arxiv}
  \vspace{-7pt}
\end{figure*}

Here, we investigate the Snowflake Hypothesis on Ogbn-Arxiv, which is representative of real-world graph scenarios. Specifically, we consider GCN, ResGCN and ResGCN+ \citep{li2020deepergcn} as backbones for evaluation. On these three backbones, we prune the adjacency matrix each layer separately guided by the cosine distance. Due to the superior ability of ResGCN and ResGCN+ to mitigate over-smoothing compared to GCN, we adopt larger values of $\rho$ on ResGCN and ResGCN+ networks. Specifically, for ResGCN and ResGCN+, we use threshold values of 0.1, 0.07, and 0.02 at layers 16, 32, and 64, respectively. For GCN, we use thresholds of 0.08, 0.05 for 16 and 32 layers. The experimental results are shown in Fig \ref{fig:arxiv}. We list observations as follows: \textbf{Obs 1.} SnoHv2, when combined with three types of backbones, can achieve the same or even better performance under relatively high sparsity conditions as compared to the original network. \textbf{Obs 2.} SnoHv1 can achieve a slightly higher improvement (Tab \ref{snoh+arxiv}) compare with SnoHv2 of citation network. Through heterogeneity analysis \citep{pei2020geom}, we find that the citation network possesses relatively more severe heterogenous networks, and the comparison at different levels might be of relatively low significance for this type of graph; we should rather avoid the aggregation of heterogenous information from the initial layers. We have placed specific analysis and conjecture in Appendix \ref{homophily}.

\begin{table}[!t]
	\centering
 \caption{Comparison between different backbones with SnoHv1 on Ogbn-Arxiv, where L denotes layers.}
 \vspace{-0.7em}
\resizebox{1\linewidth}{!}{
\begin{tabular}{ccccccccccccc}
\toprule
  \multirow{2}{*}{}&\multicolumn{3}{c}{\textbf{GCN}, $\rho=0.08, 0.05$} 
  &\multicolumn{3}{c}{\textbf{ResGCN}, $\rho=0.1, 0.07, 0.02$}
  &\multicolumn{3}{c}{\textbf{ResGCN+}, $\rho=0.1, 0.07, 0.02$}
  \\
\cmidrule(lr){2-4} \cmidrule(lr){5-7} \cmidrule(lr){8-10} 
  & \textbf{16-L} & \textbf{32-L} & \textbf{64-L} & \textbf{16-L} & \textbf{32-L} & \textbf{64-L} & \textbf{16-L} & \textbf{32-L} & \textbf{64-L}  \\
  \midrule
+SnoHv1 & 69.89  & 65.88  & Collapse  & 71.78   & 71.17  & 70.96   & 70.01  & 70.93  & 70.44    
\\

  w/o SnoHv1 & 69.47  & 63.84   & Collapse  & 70.53   & 70.43   & 70.61   & 70.79   & 70.58   & 70.17   
\\ \bottomrule
\end{tabular}
}
 \label{snoh+arxiv}
 \vspace{-1em}
\end{table}

\begin{figure*}[h]
  \centering
  \includegraphics[width=0.95\linewidth]{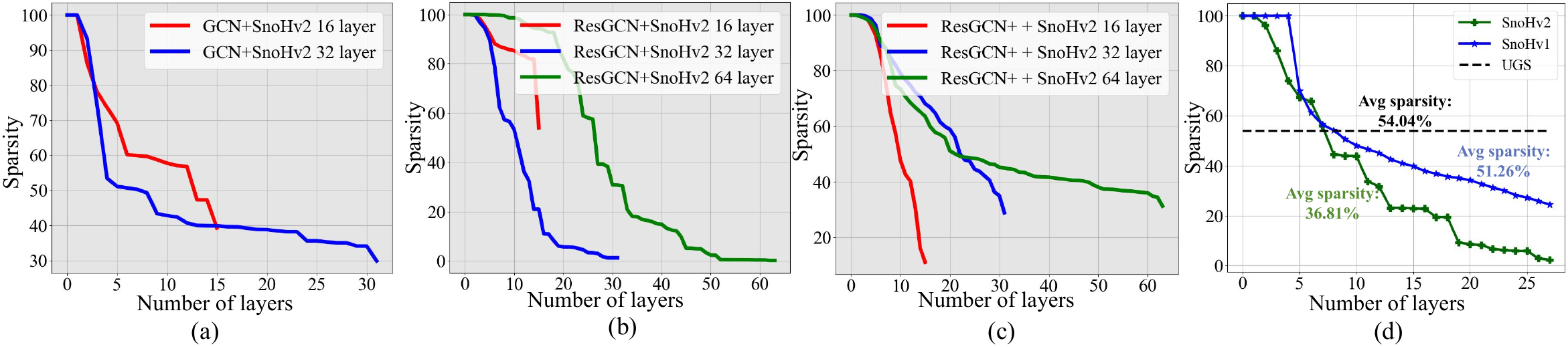}
  \vspace{-0.5em}
  \caption{(a) (b) (c) denote the edge sparsity under different backbones. It is worth noting that sparsity is represented as the ratio of the remaining edges to the total number of edges. (d) represents the sparsity of each layer under ResGCN+Arxiv setting of SnoHv1/v2 and UGS.}
  \label{fig:arxiv_spar}
  \vspace{-7pt}
\end{figure*}

Similarly, we follow UGS with a 28-layer ResGCN+Arxiv as the benchmark setting, and remove the weight pruning part. We iteratively prune at a rate of 0.05 for 20 times, observing the sparsity at which it finds the winning ticket, and compare it to the optimal sparsity of SnoHv1/v2. We control the pruning rate of each layer in SnoHv1 to be 0.3, while in SnoHv2, we set $\rho$ to 0.2. All network training for 1000 epochs with learning rate 0.001 with Adam optimizer.  As shown in Fig \ref{fig:arxiv_spar}, we find that our pruning rate on the ResGCN is higher than that of the graph lottery ticket, yet we can achieve relatively comparable performance. This corroborates the possibility that deep layer aggregation may indeed no longer contribute to the model, allowing for stopping at shallower layers.
\vspace{-1.0em}
\subsection{SnoHv2 on Ogbn-Protein and Ogbn-Product}
\vspace{-0.6em}
To verify the scalability of our model on large datasets, we further expanded the dataset size and tested its performance on datasets with tens of millions of edges. We used the Ogbn-Protein and Ogbn-Product datasets as benchmarks. Due to the high complexity of whole-graph training, we adopted the common subgraph approach \citep{DBLP:journals/corr/abs-1905-07953}. Since our pruning mask is static, we split the above two datasets into a fixed number of subgraphs for training (In this work, we set values as 30, 30 for two graphs). Based on the aforementioned algorithms, we integrated GCN, ResGCN, and ResGCN+, referring to them as Cluster-GCN, Cluster-Res and Cluster-Res+ respectively. Due to the need to measure an excessive number of edge element gradients, the implementation efficiency of SnoHv1 on these two datasets is relatively low. Therefore, we only use SnoHv2 as our method for hypothesis verification.  As shown in Tab \ref{snoh+large}, in comparison to backbones, the utilization of SnoHv2 leads to a significant enhancement in performance. These findings align with the observed behavior in smaller datasets. Specifically, under the configuration of 16 and 32 layers with the combination of protein and Cluster-Res, we managed to surpass the baseline by approximately 1.0\%. On the GCN architecture, a notable enhancement of almost 3.0\% was achieved at 32 layers. This further clarifies the validity of our hypothesis. Intriguingly, we uncovered that denser graphs, such as Ogbn-Proteins, demonstrate greater resilience to sparsification. Upon contrasting the node classification outcomes on Ogbn-ArXiv (average degree$ \approx $13.77) and Ogbn-Proteins (average degree$ \approx $597.00), it becomes evident that Ogbn-Proteins maintains only a minimal performance discrepancy with SnoHv2, even when applied to heavily pruned graphs ($ \approx $ 34.77\%, sparsity of SnoHv2+Arxix $ \approx $ 36.81\%), this finding also aligns with the conclusions drawn in \citep{chen2020lottery}.

\begin{table}[!t]
	\centering
 \caption{Comparison between different backbones with SnoHv2 on Ogbn-Proteins/Products, where L denotes layers. w/o denotes without SnoHv2.}
 \vspace{-0.7em}
\resizebox{1\linewidth}{!}{
\begin{tabular}{cccccccccccccc}
\toprule
  \multirow{2}{*}{}&\multicolumn{3}{c}{\textbf{Cluster-GCN}, $\rho=0.15$} 
  &\multicolumn{3}{c}{\textbf{Cluster-Res}, $\rho=0.15$}
  &\multicolumn{3}{c}{\textbf{Cluster-Res+}, $\rho=0.15$}
  \\
\cmidrule(lr){2-4} \cmidrule(lr){5-7} \cmidrule(lr){8-10} 
  & \textbf{16-L} & \textbf{32-L} & \textbf{64-L} & \textbf{16-L} & \textbf{32-L} & \textbf{64-L} & \textbf{16-L} & \textbf{32-L} & \textbf{64-L}  \\
  \midrule
Ogbn-Proteins  & 71.88  & 71.32  & 71.08& 79.80  & 78.87  & OOM  & 80.04  & 79.32 & OOM
\\

Ogbn-Proteins (w/o) & 71.32 & 68.44  & 70.55  & 78.40 &  77.71   & OOM  & 79.90  & 79.05  & OOM
\\ 
  \midrule
Ogbn-Product  & 68.46 & 69.44& OOM  & 79.68   & 78.99 & OOM   & 78.89  & 77.64  & OOM
\\

Ogbn-Product (w/o) & 68.40& 69.18 & OOM  &  79.50  & 78.92  & OOM  & 78.77  & 77.21  & OOM  
\\

\bottomrule
\end{tabular}
}
 \label{snoh+large}
 \vspace{-1em}
\end{table}

\vspace{-1.0em}
\section{Case Study}
\vspace{-0.7em}

\begin{wrapfigure}{r}{8cm} \vspace{-0.5em}
 \centering
 \includegraphics[width=0.56\textwidth]{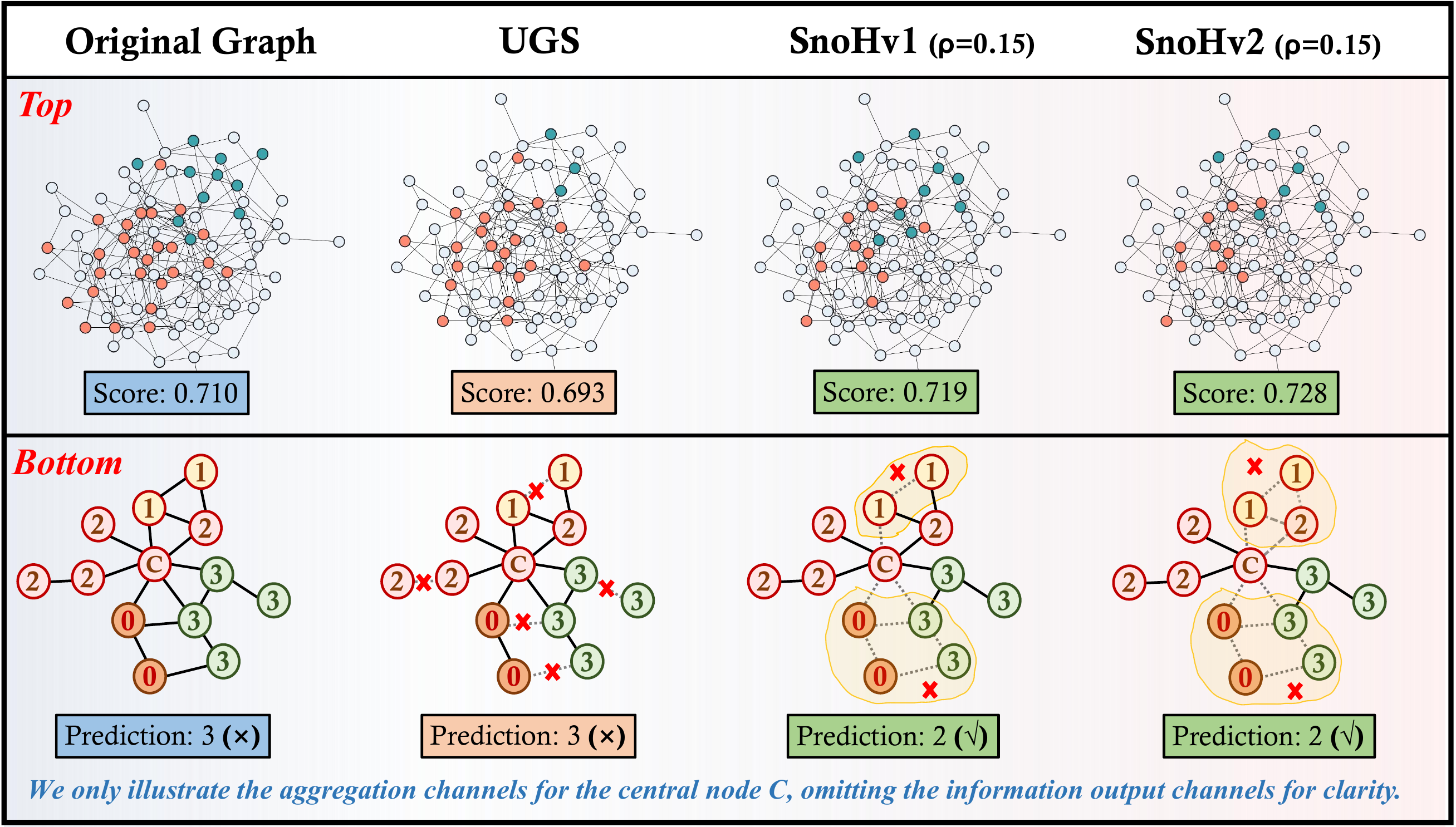}
  \vspace{-0.6em}
 \caption{\textbf{\textit{Top.}} Illustration of the two-hop receptive fields for two nodes (blue and red) along with the results showcased by different algorithms. \textbf{\textit{Bottom.}} Prediction results for the central node C (The label is 2) using different algorithms.}\label{case}
   \vspace{-1em}
\end{wrapfigure}
Although the experiments provided in Sec \ref{sec3} and \ref{sec4} are detailed, we only explore the overview performance of the SnoH. In this part, we turn to qualitatively analyze the effect of the receptive fields on some certain nodes in citepseer graph via some case studies shown in Fig \ref{case}, where all the accuracy scores belonging to the same settings and GCN baseline. Based on the information conveyed in Fig \ref{case}, the following observations can be made: \textbf{Obs 1 (top line).} Generally, we argue that nodes with a greater number of neighbors should be assigned a relatively higher pruning rate, while those with fewer neighboring information should be inclined to be retained. However, UGS does not adequately preserve the adjacency matrices of nodes with fewer neighbors within the receptive field, which can be detrimental to the prediction of certain nodes. The SnoH, in comparison to UGS, might exhibit better selectivity in this aspect. This ensures predictive capability for certain nodes and overcomes over-smoothing issues. \textbf{Obs 2 (bottom line).} Taking a microscopic look on a specific node. We observe that by blocking certain channels transmitting information to the central node, it encourages the node to pay more attention to its significant neighbors, leading to a more accurate final prediction. By reducing aggregation channels for nodes with neighbors and blocking entire aggregation channels for certain non-essential neighbors, the SnoH effectively addresses the issues of overfitting and oversmoothing.

\vspace{-1.em}
\section{Conclusion \& Future Work}
\vspace{-0.7em}
In this paper, we have proposed the Snowflake Hypothesis for the first time to discover the unique receptive field of each node, carefully suggesting the depth of early stopping for each node through the prevalent techniques of adjacency matrix pruning and cosine distance judgment. Our experiments on multiple graph datasets have demonstrated that early stopping of node aggregation at different depths can effectively enhance inference efficiency (pruning benefits), overcome the over-smoothing problem (early stopping benefits), and simultaneously offer better interpretability. Our framework is both general and succinct, compatible with many mainstream deep networks, such as ResGCN, JKNet, \textit{etc.}, to boost performance and can also be integrated with different training strategies. Our empirical study of the existence of snowflakes invites a number of future work and research questions. We have listed the potential research points and future work as follows: 

\textbf{Future work.} (1) \textit{Introducing the block concept from CV.} A relatively simple and faster way to accelerate training is to introduce the block concept from CV, combining multiple layers of adjacency matrices into one block. Within the same block, the pruning elements of all adjacency matrices are the same, and the shallow blocks align the reduced edges to the deeper layers. (2) \textit{Designating improved early stopping strategies}, in this paper we have utilized the simplest pruning strategy to determine whether a node should stop early. We anticipate that in the future, more adaptive early stopping strategies can be discovered to assist in better supporting the Snowflake Hypothesis.

\bibliography{iclr2023_conference}
\bibliographystyle{iclr2023_conference}

\clearpage
\appendix
\section{Details of SnoHv1}\label{Snov2}
\begin{figure*}[!h]
  \centering
  \includegraphics[width=1\linewidth]{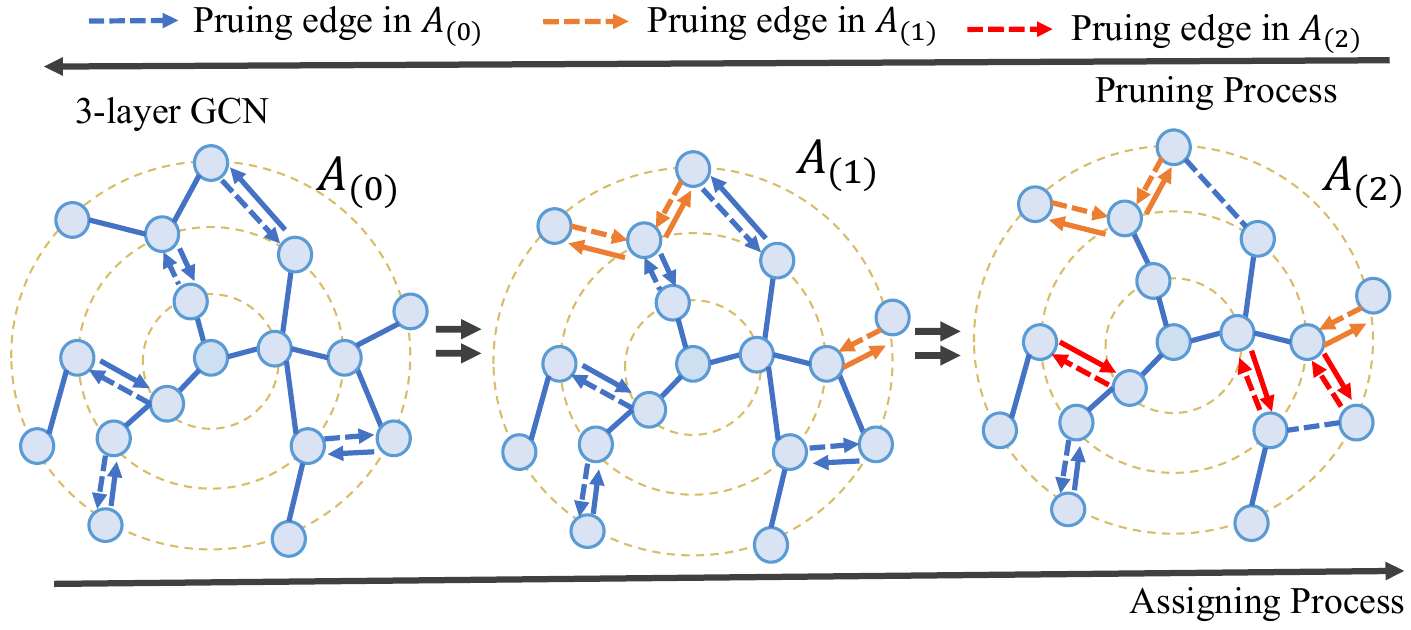}
  \caption{An example of our SnoHv1 in 3-layer GCN.}
  \label{fig:snohv2}
\end{figure*}

In this section, we start by providing a simple illustration of our second version model, SnoHv1, through examples. We then delve into the detailed description of our algorithmic workflow. As shown in the Fig \ref{fig:snohv2}, we execute our SnoHv1 on a three-layer GCN. The adjacency matrix undergoes an assigning process to align the pruning elements from inner layers with the outer layer's adjacency matrix, propagating their influence layer by layer. Additionally, the outer layer also adds pruning edges during each individual pruning, ensuring that the same node has different aggregation depths on different neighbors. Compared to SnoHv2, this type of pruning allows for more refined handling of the node's receptive field issue.

\section{Training schemes}\label{training}

\begin{figure*}[!t]
  \centering
  \includegraphics[width=1\linewidth]{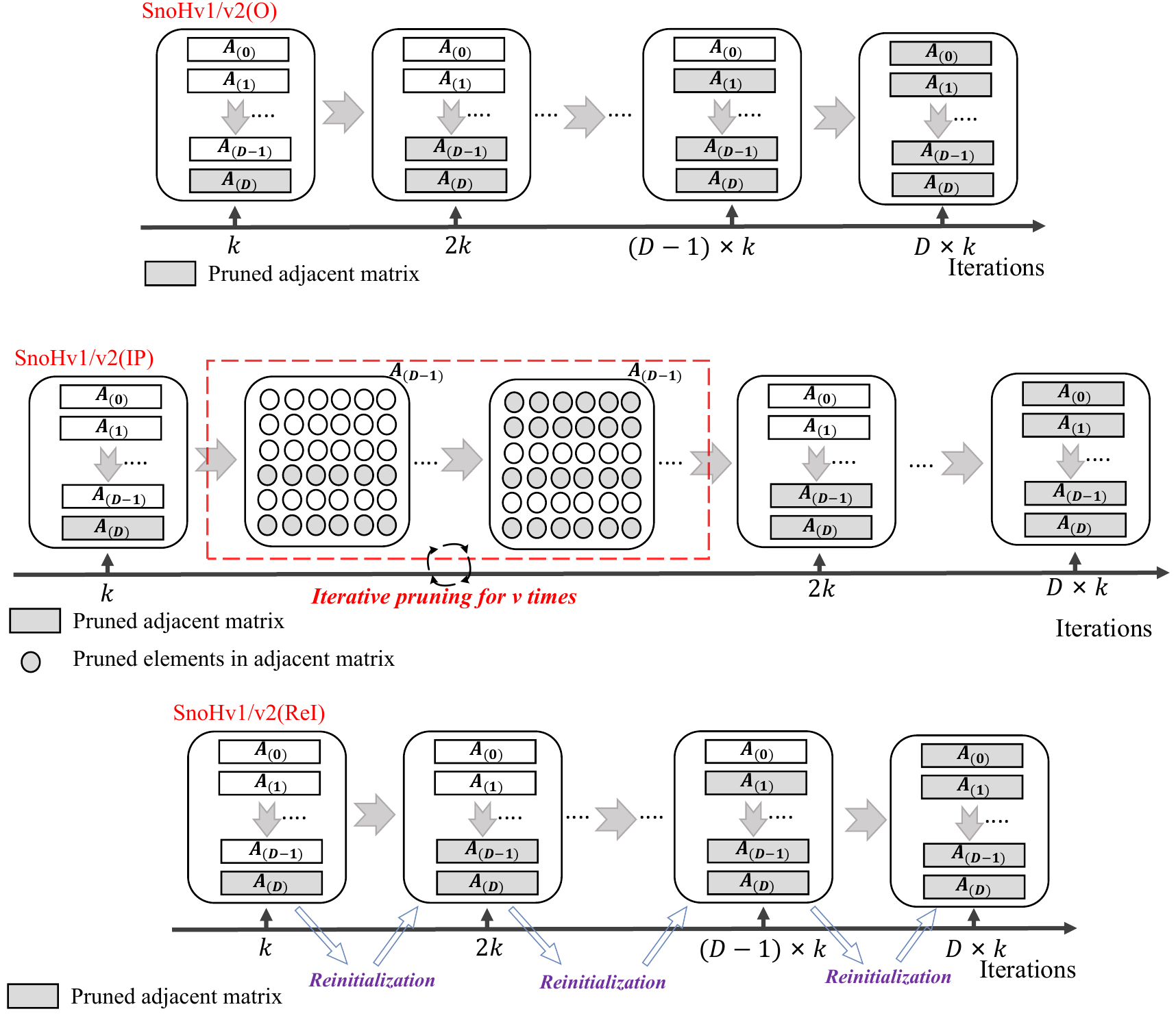}
  \caption{An illustration of three training schemes depicted in our paper.}
  \label{fig:training}
\end{figure*}

In this part, we present the detailed processes of the three training methods in the Fig \ref{fig:training}. For SnoHv1/v2(O), we employ the one-shot pruning training approach, where a complete pruning of the adjacency matrix is performed every $k$ iterations, removing p\% of the rows. This process is repeated until all adjacency matrices are traversed. For SnoHv1/v2(IP), we refine the pruning of an adjacency matrix into an iterative pruning process, where a portion of rows (v1) or elements (v2) are pruned at each regular iterations, followed by continued training and multiple pruning iterations. It is worth noting that our approach resembles the UGS algorithm \citep{chen2020lottery}, however, the key difference lies in our pruning being based on the notion of receptive fields. The inner layer's adjacency matrix influences the size of the receptive field in the outer layer, providing better interpretability and algorithmic rationality in an intuitive sense. As for SnoHv1/v2(ReI), during the training process, it is possible that while adjusting the receptive fields of the outer layer's adjacency matrix, the network parameters have already converged to a relatively good local optima. At this point, pruning the inner layer's adjacency matrix may have minimal impact on the model's performance. To address this, we employ a re-initialization (ReI) strategy. After each adjacency matrix pruning is completed, we re-initialize the entire model while keeping the pruned adjacency matrix fixed. Subsequently, we proceed to train and optimize the inner layer's adjacency matrix. Although the training processes of SnoHv1/v2(ReI) and SnoHv1/v2(O) involve $k$ iterations, they may not necessarily be the same in practice. For ease of representation, we use $k$ to denote the number of training iterations. However, in the actual implementation, we will provide specific values for the hyperparameters.

\section{Details of datasets and backbones}\label{dataset}
In this section, we provide detailed descriptions of the datasets used in this paper. The statistical characteristics of the datasets are presented in Table \ref{tab:dataset}.

\begin{table}[t]\small
  \centering
  \setlength{\tabcolsep}{4.0pt} 
  \renewcommand\arraystretch{1.2} %
  \caption{Statistical characteristics of the dataset used in our paper.}
  \vspace{-1em}
    \begin{tabular}{ccccccc}
    \toprule
    Dataset & Task  & \#Nodes & \#Edges   & \#Classes & Evaluation Metric \\
    \midrule
    Cora  & Node classification       & 2,708 & 5,278  & Multi-class      & Accuracy \\
    citepseer & Node classification        & 3,327 & 4,732  & Multi-class      & Accuracy \\
    PubMed & Node classification       & 19,717 & 88,338    & Multi-class     & Accuracy  \\ \hdashline
    Ogbn-Proteins & Node classification    & 132,534 & 39,561,252    & Binary      & ROC-AUC \\
    Ogbn-Products & Node classification   & 2,449,029 & 61,859,140  & Multi-class      & Accuracy \\ 
    Ogbn-Arxiv & Node classification   & 169,343 & 1,166,243  & Multi-class      & Accuracy \\ 
    
    \bottomrule
    \end{tabular}%
  \label{tab:dataset}%
\end{table}%

\vspace{-1em}
\section{Related Work} \label{related work}
\vspace{-0.5em}
\textbf{Graph Neural Networks}. 
The Graph Neural Networks (GNNs) family encompasses a diverse array of graph message-passing architectures, each capable of integrating topological structures and node features to create more expressive representations of the entire graph. The efficacy of GNNs, as we illustrate, primarily originates from their inherent ``message-passing'' function, represented as ${H^{\left( k \right)}} = M\left( {A,{H^{\left( {k - 1} \right)}};{{\rm{\Theta }}^{\left( k \right)}}} \right)$. Here, $H^{(k)} \in \mathbb{R}^{|\mathcal{V}| \times F}$ corresponds to the node embedding after $k$ iterations of GNN aggregation, $M$ denotes the message propagation function, and ${{\rm{\Theta }}^{\left( k \right)}}$ signifies the trainable parameters at the $k$-th layer of the GNN (${H^{\left( 0 \right)}} = X$). In light of the growing popularity of graph neural networks, a myriad of propagation functions $M$ \citep{gilmer2017neural, hamilton2017inductive} and GNN variants \citep{estrach2014spectral, velickovic2017graph, li2020deepergcn, mavromatis2020graph} have emerged.

\textbf{Deep Graph Neural Networks.} Despite the promising results obtained by GNNs, they encounter notorious over-smoothing and over-fitting issues when scaling up to deep structure. To this end, many streams of work have been dedicated to solving these issues and help GNNs have a deep structure. A prominent approach is to inherit the depth modules of CNNs to the graph realm, such as skip and residual connections \citep{li2019deepgcns, sun2019adagcn, xu2018representation, li2021deepgcns, xu2018representation, chen2020simple, xu2021optimization}. However, these works do not involve customized operations for the receptive field of each node and lack a specific understanding of graphs. Another representative is  combine deep aggregation strategies with shallow GNNs \citep{wu2019simplifying, chien2020adaptive, liu2020towards, zou2019layer, rong2019dropedge, gasteiger2019diffusion}. Similarly, these works prevent the over-smoothing issue by replacing the aggregation strategy of the entire network, lacking an understanding of node-specific differentiations. There are also some works that make efforts to theoretically propose methods for training deep GNNs \citep{xu2021optimization, min2020scattering}. However, these works are limited to specific types of GNNs, lacking generalizability and practical significance.

\textbf{Graph Pooling \& Sampling.} Graph pooling and sampling devote to reducing the computational burden of GNNs by selectively sampling sub-graphs or applying pruning methods \citep{chen2018fastgcn, eden2018provable, chen2021unified, eden2018provable, chen2021unified, gao2019graph, lee2019self}.  We divide current graph pooling or sampling techniques into two categories. (1) \emph{Sampling-based methods} aims at selecting the most expressive nodes or edges (i.e., dropping the rest) from the original graph to construct a new subgraph \citep{gao2019graph, lee2019self, ranjan2020asap, zhang2021hierarchical}. Though efficient, the dropping of nodes/edges sometimes results in severe information loss and isolated subgraphs, which may cripple the performance of GNNs \citep{wu2022structural}. (2) \emph{Clustering-based methods} learns how to cluster together nodes in the original graph, and produces a reduced graph where the clusters are set as nodes \citep{ying2018hierarchical, wu2022structural, roy2021structure}, which can remedy the aforementioned information loss problem.


\textbf{Lottery Ticket Hypothesis (LTH).} LTH articulates that a sparse and admirable subnetwork can be identified from a dense network by iterative pruning \citep{frankle2018lottery}. LTH is initially observed in dense networks and is broadly found in many fields \citep{evci2020rigging, frankle2020linear, malach2020proving, ding2021audio, chen2020lottery, chen2021unified}. Derivative theories \citep{chen2020earlybert, you2021gebt, ma2021sanity} are proposed to optimize the procedure of network sparsification and pruning. In addition to them, Dual Lottery Ticket Hypothesis (DLTH) considers a more general case to uncover the relationship between a dense network and its sparse counterparts \citep{bai2022dual, wang2022searching}. Recenlty, graph lottery ticket \citep{chen2020lottery} proposes to use iterative pruning methods on adjacency matrix and weights (called UGS approach) can obtain a graph lottery ticket during the trianing phase.

\section{Experimental settings and results on small graphs} \label{small}

\textbf{Experimental settings.} As for three small-scale graphs, we adopt the supervised node classification setting. In our implementation, we choose 60\%, 20\%, 20\% split ratio  as our train-val-test splitting of datasets. During the training phase, we choose Adam as optimizer and set learning rate as 0.01, and hidden layer deimension as 64. Tab \ref{small setting} illustrates the experimental details of three compact datasets: Cora, citepseer, and PubMed. In this table, the term ``SnoHv1(O) PE'' signifies the pruning epoch within the one-shot pruning strategy under SnoHv1, detailing the number of epochs completed prior to pruning the adjacency matrix for each layer. ``SnoHv1(ReI) PE'' signifies the epoch count for reinitialization under the reinitialization scenario. The symbol $\rho$ is indicative of the depth at which the model halts aggregation under SnoHv2, which can be interpreted as an early termination when the cosine distance between the consolidated output and the initial layer output falls beneath $\rho$. As a rule of thumb, a larger $\rho$ induces earlier termination at lesser depths. In our experiment, different depths may correspond to different values of $\rho$. We will later discuss in detail how the settings of $\rho$ values affect the performance of the model. 

\begin{table}[htbp]\small
  \centering
  \setlength{\tabcolsep}{3.5pt}
  \caption{Node sparsity (NS) and edge sparsity (ES) of each layer when GCN+SnoHv2 achieves optimal performance under three small datasets. Li denotes L-th layer.}
    \begin{tabular}{ccc|ccc}
    \toprule
    \multicolumn{6}{c}{SnoHv2 32-layer GCN $\rho$=0.001} \\
    \midrule
     \textcolor{red}{GCN+Cora}  & \textcolor{red}{GCN+citepseer} & \textcolor{red}{GCN+PubMed} & \textcolor{red}{GCN+Cora}  & \textcolor{red}{GCN+citepseer} & \textcolor{red}{GCN+PubMed} \\ \hline
    L0 NS: 100.00\% & L0 NS: 100.00\% & L0 NS: 100.00\% &     L0 ES: 100.00\% & L0 ES: 100.00\% & L0 ES: 100.00\% \\
        L1 NS: 100.00\% &     L1 NS: 100.00\% &     L1 NS: 100.00\% &     L1 ES: 100.00\% &     L1 ES: 100.00\% &     L1 ES: 100.00\% \\
        L2 NS: 66.32\% &     L2 NS: 67.06\% &     L2 NS: 94.90\% &     L2 ES: 73.86\% &     L2 ES: 78.79\% &     L2 ES: 90.66\% \\
        L3 NS: 43.24\% &     L3 NS: 41.90\% &     L3 NS: 85.85\% &     L3 ES: 50.79\% &     L3 ES: 50.60\% &     L3 ES: 78.59\% \\
        L4 NS: 30.39\% &     L4 NS: 30.69\% &     L4 NS: 74.99\% &     L4 ES: 35.32\% &     L4 ES: 39.25\% &     L4 ES: 65.41\% \\
        L5 NS: 16.47\% &     L5 NS: 20.53\% &     L5 NS: 71.90\% &     L5 ES: 22.00\% &     L5 ES: 25.37\% &     L5 ES: 61.13\% \\
        L6 NS: 14.11\% &     L6 NS: 17.79\% &     L6 NS: 67.93\% &     L6 ES: 17.49\% &     L6 ES: 22.76\% &     L6 ES: 55.95\% \\
        L7 NS: 12.63\% &     L7 NS: 17.13\% &     L7 NS: 61.69\% &     L7 ES: 15.46\% &     L7 ES: 22.29\% &     L7 ES: 48.10\% \\
        L8 NS: 9.12\% &     L8 NS: 14.70\% &     L8 NS: 59.53\% &     L8 ES: 12.25\% &     L8 ES: 18.38\% &     L8 ES: 46.21\% \\
        L9 NS: 8.42\% &     L9 NS: 14.25\% &     L9 NS: 59.23\% &     L9 ES: 11.72\% &     L9 ES: 17.99\% &     L9 ES: 45.63\% \\
        L10 NS: 8.42\% &     L10 NS: 13.98\% &     L10 NS: 54.65\% &     L10 ES: 11.72\% &     L10 ES: 17.42\% &     L10 ES: 41.52\% \\
        L11 NS: 3.21\% &     L11 NS: 10.43\% &     L11 NS: 54.31\% &     L11 ES: 6.24\% &     L11 ES: 13.64\% &     L11 ES: 40.78\% \\
        L12 NS: 3.06\% &     L12 NS: 10.10\% &     L12 NS: 54.13\% &     L12 ES: 6.19\% &     L12 ES: 13.15\% &     L12 ES: 40.65\% \\
        L13 NS: 2.95\% &     L13 NS: 9.65\% &     L13 NS: 53.77\% &     L13 ES: 6.12\% &     L13 ES: 12.45\% &     L13 ES: 40.30\% \\
        L14 NS: 2.66\% &     L14 NS: 9.56\% &     L14 NS: 52.41\% &     L14 ES: 6.00\% &     L14 ES: 12.41\% &     L14 ES: 37.90\% \\
        L15 NS: 2.66\% &     L15 NS: 7.15\% &     L15 NS: 51.70\% &     L15 ES: 6.00\% &     L15 ES: 9.87\% &     L15 ES: 37.45\% \\
        L16 NS: 2.55\% &     L16 NS: 6.88\% &     L16 NS: 49.26\% &     L16 ES: 5.90\% &     L16 ES: 9.70\% &     L16 ES: 35.97\% \\
        L17 NS: 2.29\% &     L17 NS: 6.88\% &     L17 NS: 49.09\% &     L17 ES: 5.63\% &     L17 ES: 9.70\% &     L17 ES: 35.59\% \\
        L18 NS: 2.03\% &     L18 NS: 5.77\% &     L18 NS: 48.92\% &     L18 ES: 5.49\% &     L18 ES: 8.07\% &     L18 ES: 35.36\% \\
        L19 NS: 1.88\% &     L19 NS: 5.77\% &     L19 NS: 47.60\% &     L19 ES: 5.07\% &     L19 ES: 8.07\% &     L19 ES: 34.68\% \\
        L20 NS: 1.14\% &     L20 NS: 5.53\% &     L20 NS: 47.49\% &     L20 ES: 4.23\% &     L20 ES: 7.88\% &     L20 ES: 34.58\% \\
        L21 NS: 1.00\% &     L21 NS: 5.50\% &     L21 NS: 45.56\% &     L21 ES: 3.95\% &     L21 ES: 7.83\% &     L21 ES: 32.89\% \\
        L22 NS: 0.89\% &     L22 NS: 4.84\% &     L22 NS: 44.47\% &     L22 ES: 3.89\% &     L22 ES: 6.05\% &     L22 ES: 31.81\% \\
        L23 NS: 0.89\% &     L23 NS: 4.57\% &     L23 NS: 44.02\% &     L23 ES: 3.89\% &     L23 ES: 5.81\% &     L23 ES: 31.40\% \\
        L24 NS: 0.89\% &     L24 NS: 3.91\% &     L24 NS: 43.86\% &     L24 ES: 3.89\% &     L24 ES: 5.05\% &     L24 ES: 31.28\% \\
        L25 NS: 0.89\% &     L25 NS: 3.91\% &     L25 NS: 43.53\% &     L25 ES: 3.89\% &     L25 ES: 5.05\% &     L25 ES: 30.92\% \\
        L26 NS: 0.81\% &     L26 NS: 3.85\% &     L26 NS: 43.33\% &     L26 ES: 3.46\% &     L26 ES: 5.00\% &     L26 ES: 30.82\% \\
        L27 NS: 0.55\% &     L27 NS: 3.22\% &     L27 NS: 43.28\% &     L27 ES: 2.05\% &     L27 ES: 3.23\% &     L27 ES: 30.81\% \\
        L28 NS: 0.52\% &     L28 NS: 3.01\% &     L28 NS: 41.57\% &     L28 ES: 2.04\% &     L28 ES: 2.93\% &     L28 ES: 30.25\% \\
        L29 NS: 0.52\% &     L29 NS: 2.98\% &     L29 NS: 41.40\% &     L29 ES: 2.04\% &     L29 ES: 2.91\% &     L29 ES: 30.16\% \\
        L30 NS: 0.48\% &     L30 NS: 2.98\% &     L30 NS: 41.37\% &     L30 ES: 1.99\% &     L30 ES: 2.91\% &     L30 ES: 30.15\% \\
       L31 NS: 0.48\% &     L31 NS: 2.89\% &     L31 NS: 27.80\% &     L31 ES: 1.99\% &     L31 ES: 2.76\% &     L31 ES: 17.68\% \\
    \bottomrule
    \end{tabular}%
  \label{tab:32S}%
\end{table}%

\begin{table}[htbp] \small
  \centering
  \caption{Node sparsity (NS) and edge sparsity (ES) of each layer when GCN+SnoHv2 achieves optimal performance under Cora datasets. Li denotes L-th layer.}
    \begin{tabular}{cc}
    \toprule
    \multicolumn{2}{c}{SnoHv2 64-layer GCN+Cora $\rho$=0.001} \\
    \midrule
    L17 NS: 36.93\% & L17 ES: 47.08\% \\
    L18 NS: 36.89\% & L18 ES: 47.06\% \\
    L19 NS: 36.89\% & L19 ES: 47.06\% \\
    L20 NS: 36.89\% & L20 ES: 47.06\% \\
    L21 NS: 36.41\% & L21 ES: 46.82\% \\
    L22 NS: 36.41\% & L22 ES: 46.82\% \\
    L23 NS: 36.41\% & L23 ES: 46.82\% \\
    L24 NS: 36.41\% & L24 ES: 46.82\% \\
    L25 NS: 36.37\% & L25 ES: 46.80\% \\
    L26 NS: 36.37\% & L26 ES: 46.80\% \\
    L27 NS: 36.37\% & L27 ES: 46.80\% \\
    L28 NS: 36.37\% & L28 ES: 46.80\% \\
    L29 NS: 29.21\% & L29 ES: 41.43\% \\
    L30 NS: 25.26\% & L30 ES: 37.16\% \\
    L31 NS: 25.26\% & L31 ES: 37.16\% \\
    L32 NS: 25.26\% & L32 ES: 37.16\% \\
    L33 NS: 25.26\% & L33 ES: 37.16\% \\
    L34 NS: 25.26\% & L34 ES: 37.16\% \\
    L35 NS: 25.26\% & L35 ES: 37.16\% \\
    L36 NS: 25.26\% & L36 ES: 37.16\% \\
    L37 NS: 11.82\% & L37 ES: 21.92\% \\
    L38 NS: 11.82\% & L38 ES: 21.92\% \\
    L39 NS: 7.87\% & L39 ES: 16.89\% \\
    L40 NS: 7.87\% & L40 ES: 16.89\% \\
    L41 NS: 7.87\% & L41 ES: 16.89\% \\
    L42 NS: 7.87\% & L42 ES: 16.89\% \\
    L43 NS: 7.72\% & L43 ES: 16.70\% \\
    L44 NS: 7.72\% & L44 ES: 16.70\% \\
    L45 NS: 7.53\% & L45 ES: 16.36\% \\
    L46 NS: 7.53\% & L46 ES: 16.36\% \\
    L47 NS: 7.53\% & L47 ES: 16.36\% \\
    L48 NS: 7.53\% & L48 ES: 16.36\% \\
    L49 NS: 7.53\% & L49 ES: 16.36\% \\
    L50 NS: 7.53\% & L50 ES: 16.36\% \\
    L51 NS: 7.53\% & L51 ES: 16.36\% \\
    L52 NS: 7.53\% & L52 ES: 16.36\% \\
    L53 NS: 7.53\% & L53 ES: 16.36\% \\
    L54 NS: 7.53\% & L54 ES: 16.36\% \\
    L55 NS: 7.53\% & L55 ES: 16.36\% \\
    L56 NS: 7.53\% & L56 ES: 16.36\% \\
    L57 NS: 7.53\% & L57 ES: 16.36\% \\
    L58 NS: 7.50\% & L58 ES: 16.32\% \\
    L59 NS: 6.79\% & L59 ES: 15.54\% \\
    L60 NS: 6.79\% & L60 ES: 15.54\% \\
    L61 NS: 6.79\% & L61 ES: 15.54\% \\
    L62 NS: 6.79\% & L62 ES: 15.54\% \\
    L63 NS: 6.57\% & L63 ES: 15.26\% \\
    \bottomrule
    \end{tabular}%
  \label{tab:cora64}%
\end{table}%

\begin{table*}[htbp]\small
  \centering
  \setlength{\tabcolsep}{4.0pt} 
  \renewcommand\arraystretch{1.0} %
  \caption{Implementation details of SnoH on node classification on Cora, citepseer and PubMed datasets.}
    \begin{tabular}{ccccccccc}
    \toprule
    Task  & \multicolumn{6}{c}{Node classification}  \\
    \midrule
    Dataset & Learning rate  & Optimizer & SnoHv1(O) PE  & SnoHv1(ReI) PE & $\rho $ & Total training epoch \\
    \midrule
    Cora & 0.01   & Adam   & 30    & 300  & -   & 1000 \\
    citepseer & 0.01  & Adam  & 30  & 300   & -  & 1000 \\
    PubMed  & 0.01  & Adam  & 30  & 300   & -  & 1000 \\
    Ogbn-Arxiv  & 0.001   & Adam   & 30    & -  & -   & 500 \\
    Ogbn-Protein & 0.01  & Adam  & -  & -   & -  & 75 \\
    Ogbn-Product  & 0.001  & Adam  & -  & -   & -  & 500 \\

    \bottomrule
    \end{tabular}%
  \label{small setting}%
\end{table*}%

\clearpage
\begin{figure*}[h]
  \centering
  \includegraphics[width=0.95\linewidth]{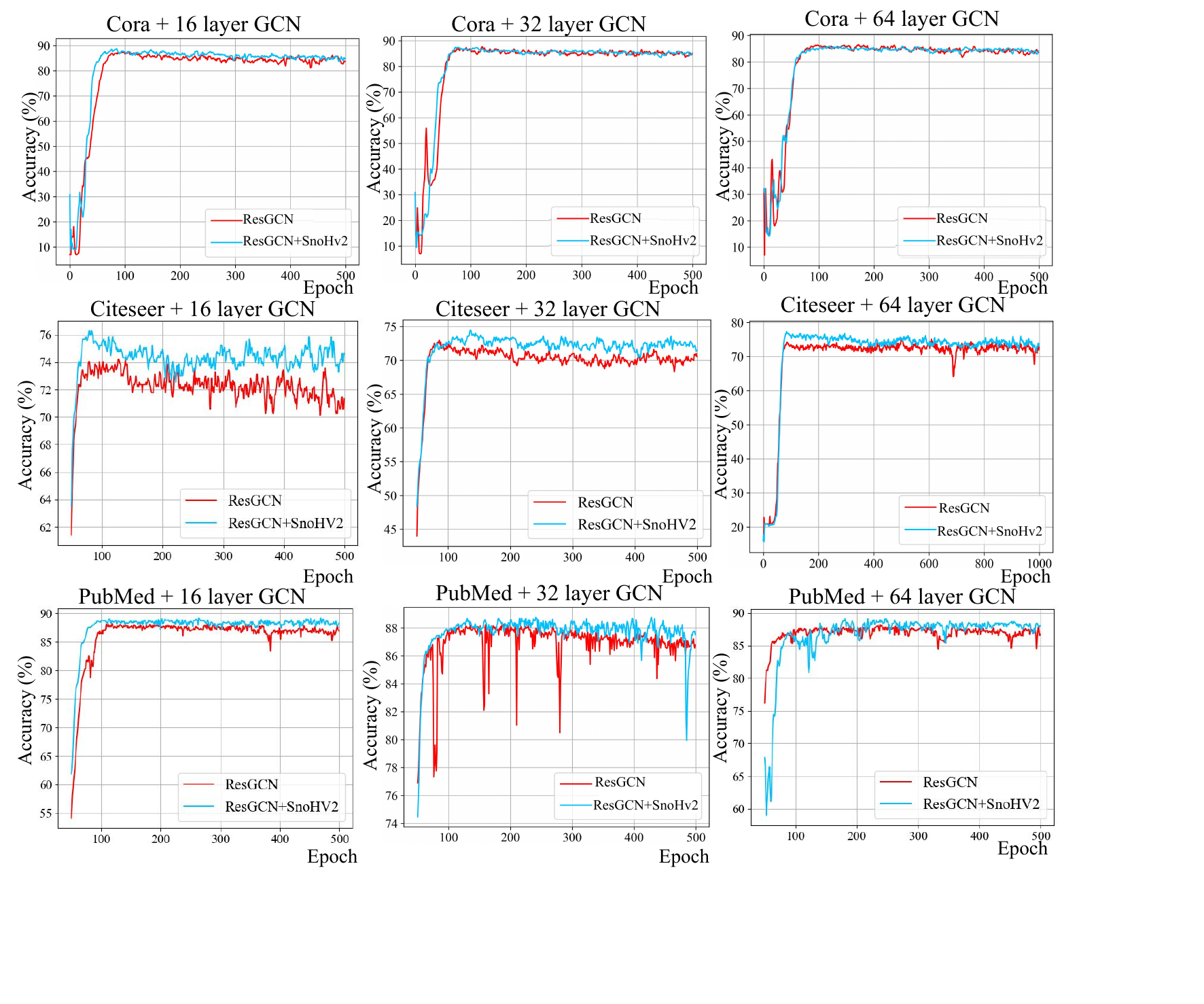}
  \vspace{-0.5em}
  \caption{Test curves during training on 16, 32, 64 layer settings using SnoHv2 across three small graphs with ResGCN backbone. We recorded the ResGCN original performance as 85.21\%, 71.65\%, 87.01\% on Cora, citepseer and PubMed at 64 layers. When combined with SnoHv2, the performance is 85.90\%, 72.94\%, 88.11\%.}
  \label{fig:res}
  \vspace{-7pt}
\end{figure*}

\begin{figure*}[h]
  \centering
  \includegraphics[width=0.95\linewidth]{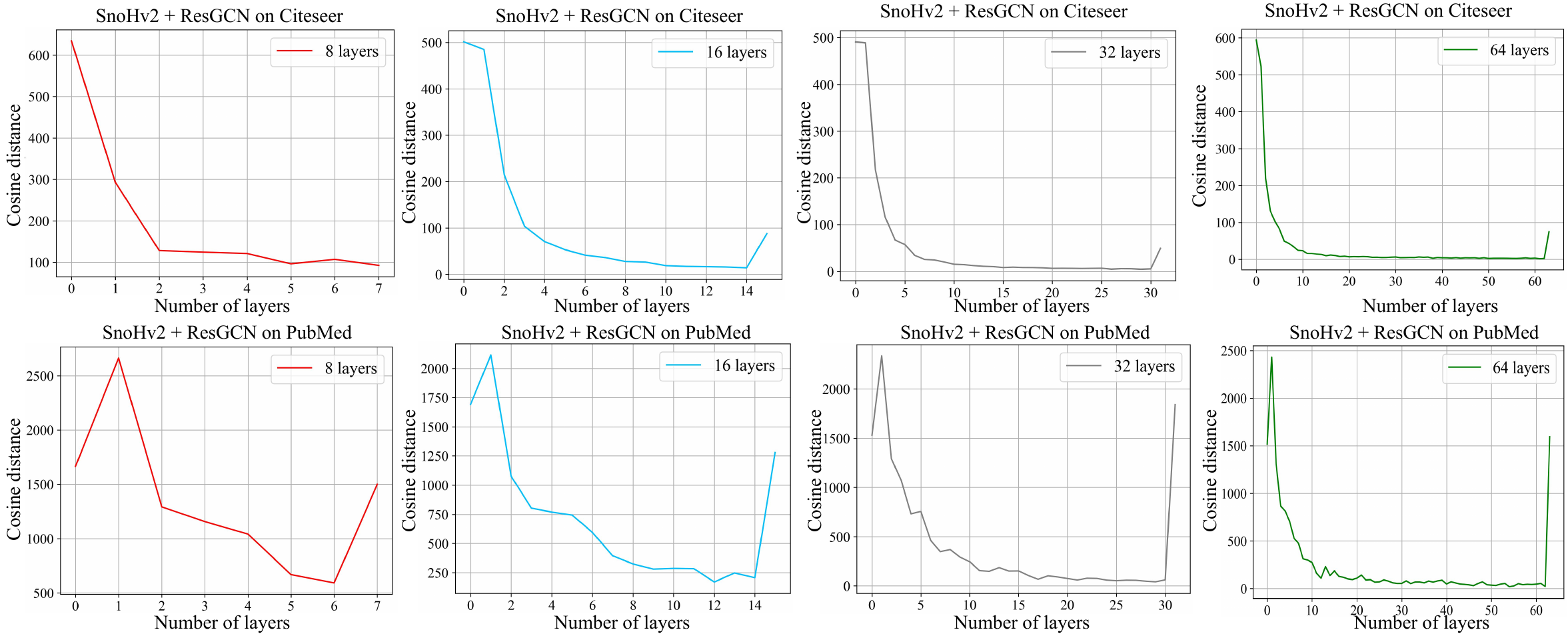}
  \vspace{-0.5em}
  \caption{The experimental settings of ResGCN+SnoHv2 on the citepseer and PubMed datasets are demonstrated using cosine distance. It can be clearly observed that a gradual decrease in cosine distance occurs across results obtained with 8 to 64 layers, indicating that as the depth of the GNN increases, the model exhibits oversmoothing phenomenon. Our approach effectively demonstrates this process.}
  \label{fig:cosine}
  \vspace{-7pt}
\end{figure*}

\clearpage
\begin{table}[htbp]\small
  \centering
    \setlength{\tabcolsep}{2.3pt}
  \caption{Edge sparsity (ES) of each layer when ResGCN+SnoHv2 or JKNet+SnoHv2 achieves optimal performance under Cora, citepseer and PubMed datasets on 32 layer settings. Li denotes L-th layer.}
    \begin{tabular}{ccc|ccc}
    \toprule
    \multicolumn{6}{c}{SnoHv2 32-L ResGCN/JKNet} \\
    \midrule
    \textcolor{red}{ResGCN+Cora}  & \textcolor{red}{ResGCN+citepseer} & \textcolor{red}{ResGCN+PubMed} & \textcolor{red}{JKNet+Cora}  & \textcolor{red}{JKNet+citepseer} & \textcolor{red}{JKNet+PubMed} \\ \hline
    L0 ES: 100.00\% & L0 ES: 100.00\% & L0 ES: 100.00\% & L0 ES: 100.00\% & L0 ES: 100.00\% & L0 ES: 100.00\% \\
    L1 ES: 92.75\% & L1 ES: 100.00\% & L1 ES: 100.00\% & L1 ES: 94.48\% & L1 ES: 89.07\% & L1 ES: 97.15\% \\
    L2 ES: 54.58\% & L2 ES: 95.17\% & L2 ES: 79.52\% & L2 ES: 80.03\% & L2 ES: 84.46\% & L2 ES: 90.99\% \\
    L3 ES: 26.89\% & L3 ES: 59.25\% & L3 ES: 49.29\% & L3 ES: 65.13\% & L3 ES: 73.92\% & L3 ES: 85.81\% \\
    L4 ES: 10.33\% & L4 ES: 10.87\% & L4 ES: 30.32\% & L4 ES: 48.10\% & L4 ES: 55.69\% & L4 ES: 74.01\% \\
    L5 ES: 5.14\% & L5 ES: 5.67\% & L5 ES: 16.31\% & L5 ES: 43.79\% & L5 ES: 37.75\% & L5 ES: 52.65\% \\
    L6 ES: 2.91\% & L6 ES: 4.57\% & L6 ES: 15.19\% & L6 ES: 24.47\% & L6 ES: 30.84\% & L6 ES: 46.98\% \\
    L7 ES: 1.95\% & L7 ES: 1.26\% & L7 ES: 10.53\% & L7 ES: 19.51\% & L7 ES: 20.73\% & L7 ES: 13.83\% \\
    L8 ES: 1.42\% & L8 ES: 0.36\% & L8 ES: 9.00\% & L8 ES: 16.45\% & L8 ES: 12.01\% & L8 ES: 13.74\% \\
    L9 ES: 0.96\% & L9 ES: 0.31\% & L9 ES: 8.92\% & L9 ES: 8.76\% & L9 ES: 5.01\% & L9 ES: 13.74\% \\
    L10 ES: 0.86\% & L10 ES: 0.31\% & L10 ES: 8.57\% & L10 ES: 8.72\% & L10 ES: 0.00\% & L10 ES: 13.74\% \\
    L11 ES: 0.62\% & L11 ES: 0.30\% & L11 ES: 8.14\% & L11 ES: 8.28\% & L11 ES: 0.00\% & L11 ES: 13.74\% \\
    L12 ES: 0.57\% & L12 ES: 0.27\% & L12 ES: 6.27\% & L12 ES: 7.78\% & L12 ES: 0.00\% & L12 ES: 13.74\% \\
    L13 ES: 0.50\% & L13 ES: 0.26\% & L13 ES: 5.56\% & L13 ES: 0.00\% & L13 ES: 0.00\% & L13 ES: 0.00\% \\
    L14 ES: 0.46\% & L14 ES: 0.18\% & L14 ES: 4.14\% & L14 ES: 0.00\% & L14 ES: 0.00\% & L14 ES: 0.00\% \\
    L15 ES: 0.36\% & L15 ES: 0.15\% & L15 ES: 2.67\% & L15 ES: 0.00\% & L15 ES: 0.00\% & L15 ES: 0.00\% \\
    L16 ES: 0.30\% & L16 ES: 0.14\% & L16 ES: 2.38\% & L16 ES: 0.00\% & L16 ES: 0.00\% & L16 ES: 0.00\% \\
    L17 ES: 0.28\% & L17 ES: 0.04\% & L17 ES: 2.36\% & L17 ES: 0.00\% & L17 ES: 0.00\% & L17 ES: 0.00\% \\
    L18 ES: 0.27\% & L18 ES: 0.04\% & L18 ES: 2.22\% & L18 ES: 0.00\% & L18 ES: 0.00\% & L18 ES: 0.00\% \\
    L19 ES: 0.09\% & L19 ES: 0.03\% & L19 ES: 1.65\% & L19 ES: 0.00\% & L19 ES: 0.00\% & L19 ES: 0.00\% \\
    L20 ES: 0.09\% & L20 ES: 0.03\% & L20 ES: 1.65\% & L20 ES: 0.00\% & L20 ES: 0.00\% & L20 ES: 0.00\% \\
    L21 ES: 0.09\% & L21 ES: 0.03\% & L21 ES: 1.58\% & L21 ES: 0.00\% & L21 ES: 0.00\% & L21 ES: 0.00\% \\
    L22 ES: 0.09\% & L22 ES: 0.01\% & L22 ES: 1.55\% & L22 ES: 0.00\% & L22 ES: 0.00\% & L22 ES: 0.00\% \\
    L23 ES: 0.07\% & L23 ES: 0.01\% & L23 ES: 1.34\% & L23 ES: 0.00\% & L23 ES: 0.00\% & L23 ES: 0.00\% \\
    L24 ES: 0.07\% & L24 ES: 0.01\% & L24 ES: 1.22\% & L24 ES: 0.00\% & L24 ES: 0.00\% & L24 ES: 0.00\% \\
    L25 ES: 0.07\% & L25 ES: 0.01\% & L25 ES: 1.08\% & L25 ES: 0.00\% & L25 ES: 0.00\% & L25 ES: 0.00\% \\
    L26 ES: 0.07\% & L26 ES: 0.00\% & L26 ES: 0.90\% & L26 ES: 0.00\% & L26 ES: 0.00\% & L26 ES: 0.00\% \\
    L27 ES: 0.07\% & L27 ES: 0.00\% & L27 ES: 0.90\% & L27 ES: 0.00\% & L27 ES: 0.00\% & L27 ES: 0.00\% \\
    L28 ES: 0.05\% & L28 ES: 0.00\% & L28 ES: 0.80\% & L28 ES: 0.00\% & L28 ES: 0.00\% & L28 ES: 0.00\% \\
    L29 ES: 0.04\% & L29 ES: 0.00\% & L29 ES: 0.76\% & L29 ES: 0.00\% & L29 ES: 0.00\% & L29 ES: 0.00\% \\
    L30 ES: 0.04\% & L30 ES: 0.00\% & L30 ES: 0.61\% & L30 ES: 0.00\% & L30 ES: 0.00\% & L30 ES: 0.00\% \\
    L31 ES: 0.04\% & L31 ES: 0.00\% & L31 ES: 0.56\% & L31 ES: 0.00\% & L31 ES: 0.00\% & L31 ES: 0.00\% \\
    \bottomrule
    \end{tabular}%
  \label{tab:resjk}%
\end{table}%

\begin{figure*}[h]
  \centering
  \includegraphics[width=0.95\linewidth]{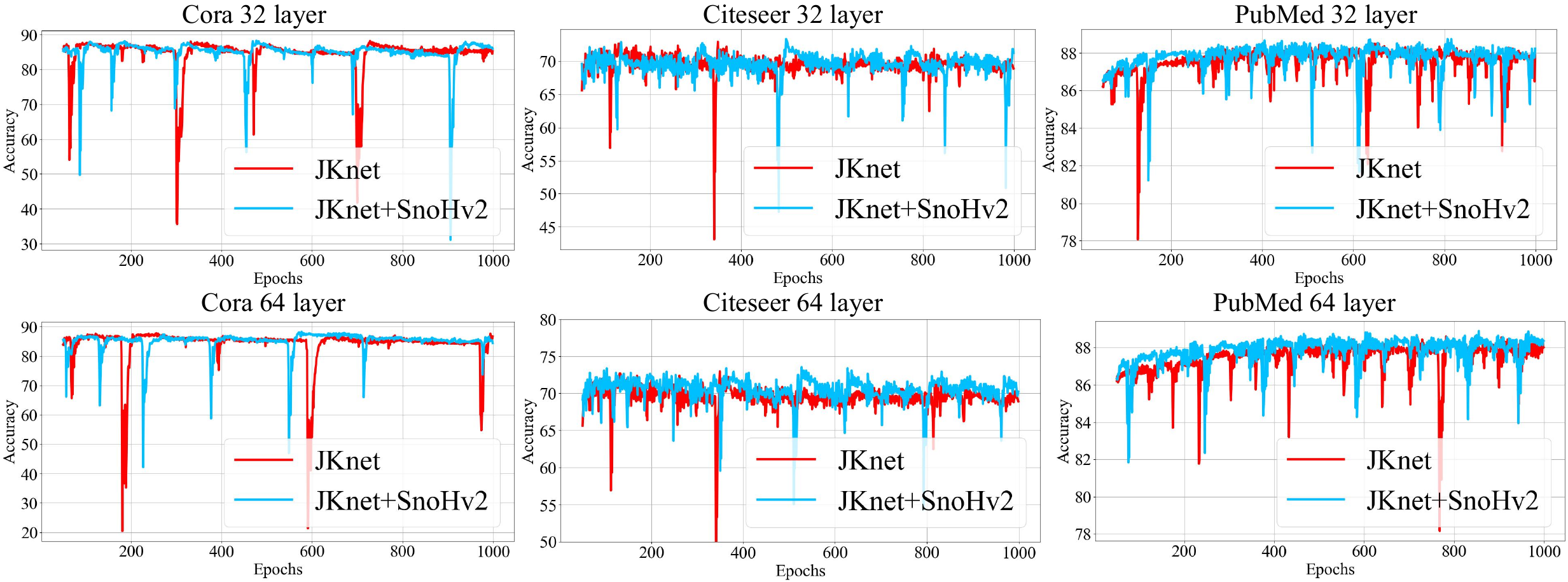}
  \vspace{-0.5em}
  \caption{The experimental settings of JKNet+SnoHv2 on the Cora, citepseer and PubMed datasets with 32 and 64 layers. }
  \label{fig:JKNet}
  \vspace{-7pt}
\end{figure*}

\clearpage
\textbf{Training scheme on SnoHv1.} In this section, we further test different training strategies for SnoHv1, including one-shot pruning, iterative pruning, and re-initialization pruning strategies. (1) The commonly used strategy is one-shot pruning, where we prune the adjacency matrix of each layer during each training process. (2) The iterative pruning method involves splitting each training process and pruning some elements of the adjacency matrix in each layer during each epoch. The training continues iteratively by removing elements from the adjacency matrix. (3) The re-initialization strategy prunes one layer of the adjacency matrix at a time. We set the training epoch to 200, meaning that every 200 epochs, we determine which elements in each layer's adjacency matrix should be pruned. 

We found that under different training strategies, there was no significant difference in the model's performance. All three training strategies achieved similar performance levels. However, the iterative pruning process, which involves repeatedly determining important parameters, was executed on the CPU and proved challenging to accelerate using a GPU. Additionally, when applied to large graphs, this iterative parameter evaluation process consumed a substantial amount of time. Similarly, the re-initialization method, with its repeated training and reinitialization to assess important parameters in each layer, resulted in significant time wastage. In some cases, it even took more than $D$ times the original training time ($D$ is the network depth for GNNs), which hampers its scalability to large graphs. \textbf{Based on the above observations, we conducted tests on our framework using large graphs, specifically employing the SnoHv2 version. We believe that our findings can provide valuable insights for future research in evaluating and testing various new designs.}

\begin{table}[t] \scriptsize
\caption{Performance comparisons on 8, 16, 32 layer settings using SnoHv1(O), SnoHv1(IP), and SnoHv1(ReI) across three small graphs, all experimental results are the average of three runs.}\label{tab:train scheme}
\vspace{-1.0em}
\setlength{\tabcolsep}{2pt}
\begin{center}
\def \arraystretch{0.98}
 \begin{tabular}{ccccccccccccc} 
 \toprule
    \multirow{2}{*}{\bf Backbone}  & \multicolumn{3}{c}{\bf 8 layers} & \multicolumn{3}{c}{\bf 16 layers}  &  \multicolumn{3}{c}{\bf 32 layers}   \\ 
    \cmidrule(l){2-4} \cmidrule(l){5-7} \cmidrule(l){8-10}  
    
     & \scriptsize \bf SnoHv1(O) & \scriptsize \bf SnoHv1(IP)  & \scriptsize \bf SnoHv1(ReI) & \scriptsize \bf SnoHv1(O) & \scriptsize \bf SnoHv1(IP)  & \scriptsize \bf SnoHv1(ReI)
     
     & \scriptsize \bf SnoHv1(O) & \scriptsize \bf SnoHv1(IP)  & \scriptsize \bf SnoHv1(ReI)    \\ 
     
     \midrule
     
       \multicolumn{9}{l}{\scriptsize{  \demph{ \it{Train scheme: SnoHv1(O), Dataset: Cora, 2-layer performance: GCN without BN = 85.37 }}}}\\
       
        GCN    & 84.37 & 83.77  & 84.30   & 83.19 & 82.21 & 82.77  & 82.09 & 82.76  & 82.45 \\
        ResGCN    & 85.12 & 85.47  & 84.99    & 84.35 & 85.17 & 84.72 & 85.37 & 85.21 & 85.87 \\
        JKNet    & 85.43 & 85.35  & 84.44    & 86.11 & 85.87 & 86.01     & 86.47 & 87.39  & 85.21 \\ 
    
    \midrule
           \multicolumn{9}{l}{\scriptsize{  \demph{ \it{Train scheme: SnoHv1/v2(O),  Dataset: citepseer, 2-layer performance: GCN without BN = 72.44 } }} }\\
       
        GCN    & 73.39 & 73.45  & 72.11   & 72.29 & 72.45 & 72.10  & 68.71 & 67.59  & 68.98 \\
        
        ResGCN    & 71.71 & 71.54  & 72.01  & 71.98 & 70.14 & 70.43   & 72.01 & 71.45  & 70.49 \\
        
        JKNet    & 71.34 & 70.57  & 71.38    & 70.47 & 70.98 & 71.03    & 69.89 & 68.47  & 68.47  \\ 
    
    \midrule
           \multicolumn{9}{l}{\scriptsize{  \demph{ \it{Train scheme: SnoHv1/v2(O), Dataset: PubMed, 2-layer performance: GCN without BN = 86.50 } }} }\\
       
        GCN    & 86.15 & 85.78  & 86.21    & 84.37 & 84.23 & 84.65 & 83.54 & 83.66 & 83.06 \\
        ResGCN    & 87.41 & 86.24  & 85.30    & 87.40 & 86.54 & 86.01 & 87.26 & 87.01  & 86.45 \\
        
        JKNet    & 88.29 & 87.68  & 88.11    & 87.27 & 86.57 & 85.92   & 88.65  & 87.81  & 86.53 \\ 
    
    \midrule

\end{tabular}
\end{center}
\vspace{-6mm}
\end{table}

\textbf{The effect of $\rho$ on SnoHv2.} Interestingly, we observed varying sensitivities of the parameter $\rho$ across different datasets. The extreme sparsity of the deep adjacency matrix depends on the properties of graphs and backbones. Specifically, on sparse graphs like Cora and citepseer, as the depth of the GCN increases, the stop rate $\rho$ gradually decreases. For example, with a 16-layer Cora+SnoHv2 configuration, the optimal value for $\rho$ is 0.4, while for 32 and 64 layers, the optimal values are 0.2 and 0.05, respectively. However, on moderately large datasets such as PubMed, sometimes a larger value of $\rho$ can lead to performance improvement.

This phenomenon also shows slight variations with different backbone architectures. When introducing residual structures, the sparsity of the deep adjacency matrix becomes even higher. This might be because residual structures preserve shallow layer information, reducing the need for deep layer information to assist in predictions.

\clearpage
\textbf{Compare with graph lottery tickets (UGS algorithm).} In our experiment, we compared the model performance of GCN+SnoHv2 and UGS, as well as random pruning under configurations of 8, 16, and 32 layers. We were pleasantly surprised to find that our results considerably outperformed those of random pruning and UGS, particularly under these deep-layer conditions. Our model demonstrated excellent performance. For instance, under a 16-layer setup for the citepseer dataset, our results were 6.57\% better than the best UGS configuration (five iterations of pruning, with each iteration pruning 5\%). This trend was consistent across all datasets and under various depth settings, further corroborating the superior capabilities of our algorithm in deep scenarios.

Upon further analysis, we believe that our enhanced performance stems from the early stopping of the receptive field for some nodes in graphs. In fact, our network can be understood as a GCN in the shallow layers and approximates an MLP in the deeper layers. While it ceases to aggregate information for nodes in the deeper layers, it successfully circumvents the issue of gradient vanishing that often plagues deep MLPs.

\begin{table}[h] 
\small
\setlength{\tabcolsep}{2.4pt}
  \caption{Comparison performances of SnoHv2 with UGS and random pruning (RP). Here IPR denotes iterative pruning rate and we set number of layers as 8. We use GCN backbone and set early stopping threshold of cosine distance as $\rho$ (Detailed descriptions in Appendix \ref{small}).} \label{UGSapp}
  \vspace{-0.6em}
  \centering
  \begin{tabular}{ccccccc}
    \toprule
     \textbf{Dataset}   &   \textbf{RP} &  \textbf{UGS(IPR=5\%)}  &  \textbf{UGS(IPR=10\%)}  & \textbf{UGS(IPR=20\%)} &  \textbf{SnoHv2} &  \textbf{GCN}  \\
    \midrule
      Cora (L=8) & $69.60$ & $73.64$ & $66.01$ & $53.29$ & \cellcolor{gray!30}${85.68}$  & ${85.11}$ \\
      citepseer (L=8) & $45.50$ & $65.80$  & ${51.50}$ & ${43.10}$ & \cellcolor{gray!30}${73.24}$ & ${72.39}$   \\
      PubMed (L=8) & $77.82$ & $84.33$   & $80.91$ & $71.05$ &\cellcolor{gray!30} ${86.56}$ & ${86.41}$   \\ \hdashline

      Cora (L=16) & $51.98$ & $60.32$ & $55.53$ & $47.24$ & \cellcolor{gray!30}${84.19}$  & ${83.75}$ \\
      citepseer (L=16) & $60.36$ & $66.31$  & ${58.12}$ & ${30.13}$ & \cellcolor{gray!30}${72.33}$ & ${71.28}$   \\
      PubMed (L=16) & $53.22$ & $79.22$   & $72.52$ & $58.39$ &\cellcolor{gray!30} ${85.79}$ & ${84.77}$   \\  \hdashline

      Cora (L=32) & $58.25$ & $69.25$ & $53.64$ & $39.20$ & \cellcolor{gray!30}${83.09}$  & ${80.33}$ \\
      citepseer (L=32) & $51.95$ & $57.37$  & $50.31$ & $51.24$ & \cellcolor{gray!30}${69.89}$ & ${68.99}$   \\
      PubMed (L=32) & $58.32$ & $77.42$   & $64.26$ & $60.77$ &\cellcolor{gray!30} ${84.06}$ & ${83.76}$   \\   

    \bottomrule
  \end{tabular}
  \vspace{-1em}
\end{table}

\section{Generalization Validation Experiments on GIN and GAT}\label{genapp}

\begin{figure*}[h]
  \centering
  \includegraphics[width=0.95\linewidth]{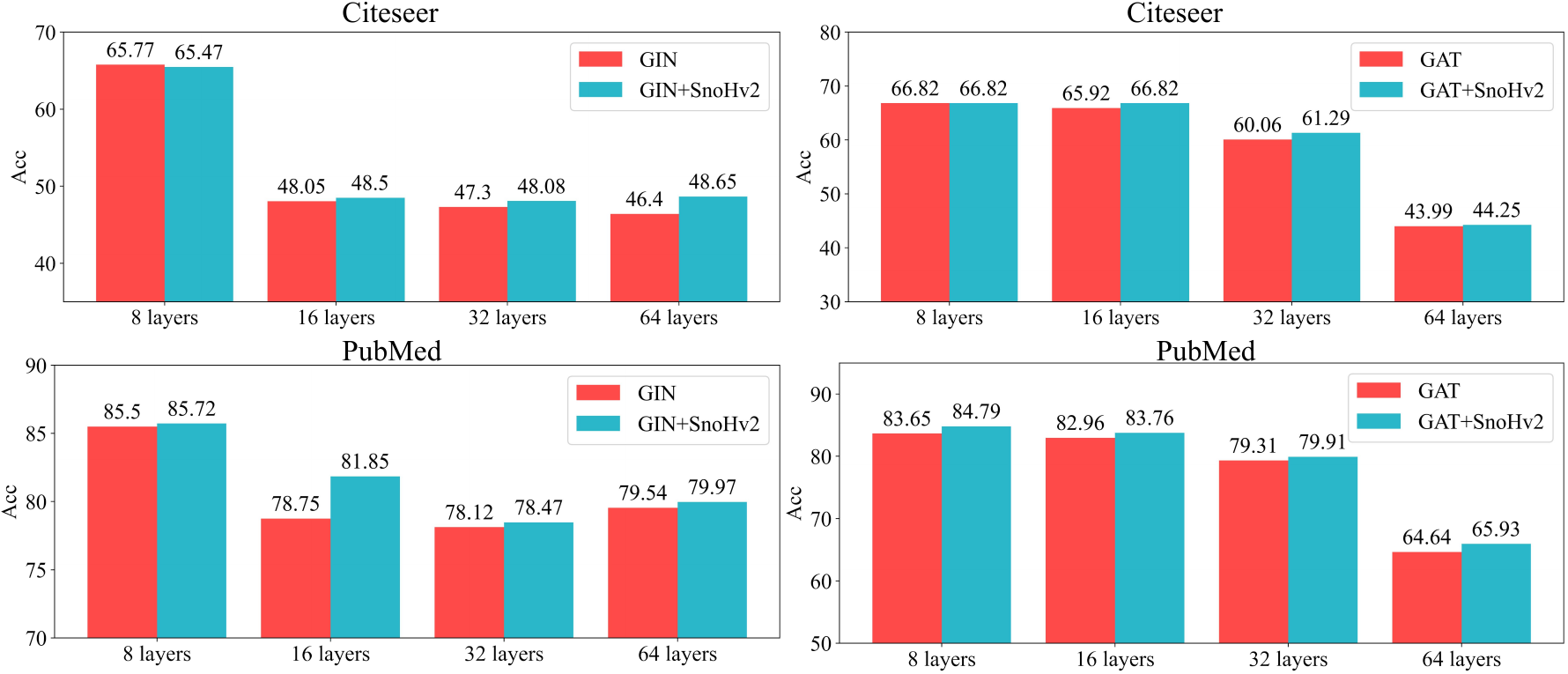}
  \vspace{-0.5em}
  \caption{The experimental settings of GIN+SnoHv2 and GAT+SnoHv2 on the citepseer and PubMed datasets are demonstrated using cosine distance. It can be readily observed that our algorithm significantly improves the performance of various GNN backbones. }
  \label{fig:cosineapp}
  \vspace{-7pt}
\end{figure*}

\clearpage
\section{Homophily ratio} \label{homophily}
\textbf{Definition.} Homophily indicates that adjacent nodes in the graph are likely to have similar attributes or labels. In a social network, for example, people with similar interests or beliefs tend to connect with each other. This pattern holds true in various kinds of networks, and its presence can significantly affect the way GNNs process and learn from the graph.

\textbf{Impact on GNN Learning.} In GNNs, information is often propagated between neighboring nodes, and node embeddings are updated based on the features of adjacent nodes. If the graph exhibits homophily, this propagation of information is likely to reinforce consistent features among neighboring nodes, which can make learning tasks like node classification more tractable.

\textbf{Challenges.} Conversely, if a graph does not exhibit homophily (i.e., similar nodes are not more likely to be connected), this can present challenges for learning. GNN models might have difficulty making accurate predictions or inferences in such cases, as neighboring nodes may provide conflicting or less relevant information.

\textbf{Measuring Homophily.} In some scenarios, quantifying the level of homophily can be beneficial for understanding the graph's structure and for selecting or designing appropriate models or algorithms. Various metrics and analyses might be used to gauge the extent of homophily within a given graph.

\textbf{Heterophily.} The opposite of homophily is heterophily, where neighboring nodes are more likely to be dissimilar. Recognizing whether a graph is more homophilous or heterophilous can be essential in choosing the correct approach and model for graph-based learning tasks.

In summary, homophily within GNNs signifies the inclination of connected nodes to exhibit similar attributes. This phenomenon is fundamental to the way GNNs interpret and learn from graphs, guiding not only the design but also the interpretation of various graph learning tasks. Its understanding leads to more effective model development and nuanced analysis.

\begin{equation}\small \label{eq1}
\frac{1}{|\mathcal{V}|} \sum_{v \in \mathcal{V}} \frac{\left|\left\{(w, v): w \in \mathcal{N}(v) \wedge y_v=y_w\right\}\right|}{|\mathcal{N}(v)|}
\end{equation} 

Through heterogeneity analysis \citep{pei2020geom}, We use Eq \ref{eq1} to calculate the degree of isomorphism in Arxiv, and we find that the homophily degree in Arxiv is relatively low (0.635).  This might cause our SnoHv2 to be deeper in the early stopping networks when judging the cosine distance at the hierarchical layer, without overcoming the problem of early aggregation.  As a result, this may lead to an insignificant improvement in our SnoHv2. As shown in Figure 5, we find that our pruning rate on the 28-layer resgcn is higher than that of the lottery ticket, yet we can achieve relatively comparable performance. This corroborates the possibility that low-level aggregation may indeed no longer contribute to the model, allowing for early stopping at shallower layers.

\end{document}